%% file: main.tex
\documentclass[11pt]{article}

\usepackage[preprint]{acl}

\usepackage{times}
\usepackage{latexsym}

\usepackage[T1]{fontenc}

\usepackage[utf8]{inputenc}

\usepackage{microtype}

\usepackage{inconsolata}

\usepackage{graphicx}
\usepackage{hyperref}
\usepackage{url}
\usepackage{graphicx} 
\usepackage{tcolorbox}
\usepackage{caption}
\usepackage{subfigure}
\usepackage{booktabs} 
\usepackage{wrapfig}
\usepackage{enumitem}

\usepackage{tabularx}
\usepackage{booktabs} 
\usepackage{amsmath} 
\usepackage{geometry} 
\usepackage{xcolor, soul}
\usepackage{pifont}
\usepackage{xspace}
\usepackage{multirow}

\input{_define}
\input{math_commands.tex}

\title{Can Post-Training Transform LLMs into Causal Reasoners?}

\author{
  \textbf{Junqi Chen$^{1,2}$ \hspace{1em} Sirui Chen$^{1,3}$ \hspace{1em} Chaochao Lu$^{1}$\thanks{Corresponding author.}}  \\
 $^1$Shanghai Artificial Intelligence Laboratory, $^2$Fudan University, $^3$Tongji University\\
\texttt{ \{chenjunqi, chensirui, luchaochao\}@pjlab.org.cn}}

\begin{document}

\maketitle

\input{sections/0_abstract}
\input{sections/1_introduction}
\input{sections/3_methodology}

\input{sections/4_experiment}
\input{sections/5_related_works}
\input{sections/6_conclusion}
\input{sections/9_ethical}
\input{sections/8_limitation}
\input{sections/10_acknowledgments}

\bibliography{acl2026_conference}

\clearpage
\appendix

\input{sections/7_appendix}

\end{document}

%% file: _define.tex
\newcommand{\sys}{\texttt{CauGym}\xspace}

\newcommand{\base}{DeepSeek-R1-Distill-Qwen-14B}
\newcommand{\ds}{DeepSeek-R1-0528-671B}
\newcommand{\reph}{\sys-rephrased}
\newcommand{\tf}{\sys-omitted}
\newcommand{\hard}{\sys-deconfounding}
\newcommand{\redun}{\sys-redundant}
\newcommand{\lack}{\sys-insufficient}
\newcommand{\qwen}{Qwen3-235B}
\newcommand{\llama}{Llama-3.3-70B}
\newcommand{\gemini}{Gemini 2.5 Pro}
\newcommand{\openai}{OpenAI o3}

\usepackage{xcolor}


%% file: math_commands.tex
\usepackage{amsmath,amsfonts,bm}

\def\eqref#1{equation~\ref{#1}}

\def\1{\bm{1}}

\DeclareMathAlphabet{\mathsfit}{\encodingdefault}{\sfdefault}{m}{sl}
\SetMathAlphabet{\mathsfit}{bold}{\encodingdefault}{\sfdefault}{bx}{n}



%% file: sections/0_abstract.tex
\begin{abstract}
Causal inference is essential for decision-making but remains challenging for non-experts. While large language models (LLMs) show promise in this domain, their precise causal estimation capabilities are still limited, and the impact of post-training on these abilities is insufficiently explored. This paper examines the extent to which post-training can enhance LLMs’ capacity for causal inference.
We introduce \sys, a comprehensive dataset comprising seven core causal tasks for training and five diverse test sets. Using this dataset, we systematically evaluate five post-training approaches: SFT, DPO, KTO, PPO, and GRPO. Across five in-domain and four existing benchmarks, our experiments demonstrate that appropriate post-training enables smaller LLMs to perform causal inference competitively, often surpassing much larger models. Our 14B-parameter model achieves 93.5\% accuracy on the CaLM benchmark, compared to 55.4\% by \openai. Furthermore, the post-trained LLMs exhibit strong generalization and robustness under real-world conditions such as distribution shifts and noisy data. Collectively, these findings provide the first systematic evidence that targeted post-training can produce reliable and robust LLM-based causal reasoners. 
Our data and GRPO-model are available at \texttt{\small \url{ https://github.com/OpenCausaLab/CauGym}}.

\end{abstract}

%% file: sections/1_introduction.tex
\section{Introduction}

Causal inference, a core component of human cognition, seeks to distinguish causation from association by estimating their effects between variables \citep{pearl2009causality,sloman2009causal}. Causal inference is crucial because decision-makers must both predict intervention effects and evaluate counterfactual outcomes \citep{woodward2005making,shpitser2006identification,bunge2017causality,chen-etal-2024-clear}. For example, one may estimate how deploying a treatment changes population health and what the same patients’ outcomes would have been had they not been treated \citep{pearl2018book}.

Many statistical methods have been developed for causal inference with observational data \citep{pmlr-v6-pearl10a}. Broadly, these methods recover causal effects either by approximating randomized assignment via adjustment for measured confounders or by exploiting quasi-experimental variation that makes treatment as-if random \citep{rubin1974estimating,rubin2005causal,pearl2016causal}. To facilitate the application of these methods, there has been a surge in the development of new causal inference libraries \citep{econml,dowhy,chen2020causalml}. They encapsulate complex algorithms, providing researchers with systematic tools for analysis and lowering the barrier to applying causal inference.
Despite lowering entry barriers, these libraries remain difficult to use correctly for non-experts. One must still articulate an identification strategy, verify assumptions, and interpret diagnostics. This challenge naturally leads to the question of whether we can develop a \emph{causal reasoner} that explains its assumptions and reasoning in plain language, making the causal inference process fully auditable.

LLMs appear promising for addressing this challenge. Their natural-language interfaces can help non-experts articulate identification questions, surface assumptions, and obtain step-by-step explanations of analyses. They have shown striking performance on tasks requiring complex reasoning, e.g., mathematics \citep{luo2025wizardmath}, coding \citep{nam2024using}, and formal theorem proving \citep{quan2024verification}.
Despite these strengths, studies show that LLMs still struggle with causal inference—especially when precise numerical estimation is required \citep{jin2023cladder,chen2024causal,jin2024can}. Moreover, some work suggests that LLMs may be inherently incapable of performing formal causal reasoning \citep{chi2024unveiling}. Although various post-training methods have proven effective in enhancing the reasoning capabilities of LLMs \citep{wang2024math,guan2025rstarmath,guo2025deepseek}, no systematic research has yet explored whether---and to what extent---these gains transfer to causal inference. Therefore, this paper addresses this gap by asking:
\begin{tcolorbox}[colback=teal!10!white, colframe=gray!70!white]
    \centering\emph{Can LLMs become effective causal reasoners through post-training?}
\end{tcolorbox}
To address this question,  a training corpus was constructed to cover seven causal inference tasks \citep{rubin2005causal,pearl2016causal}: average treatment effect (ATE), controlled direct effect (CDE), effect of the treatment on the treated (ETT), natural direct effect (NDE), natural indirect effect (NIE), probability of necessity (PN), and probability of sufficiency (PS). Together, these tasks span both interventions and counterfactuals, enabling a comprehensive strengthening of an LLM’s causal inference capabilities \citep{chen2024causal,chen-etal-2024-cello}.

Then, three categories of post-training methods are evaluated: supervised fine-tuning (SFT), offline reinforcement learning (RL), and online RL. These categories encompass five representative algorithms: SFT includes vanilla SFT \citep{wei2022finetuned}; offline RL includes Direct Preference Optimization (DPO) \citep{rafailov2023direct} and Kahneman–Tversky Optimization (KTO) \citep{ethayarajh2024kto}; online RL includes Proximal Policy Optimization (PPO) \citep{schulman2017proximal,ouyang2022training} and Group Relative Policy Optimization (GRPO) \citep{shao2024deepseekmath}.

Finally,  the LLMs are evaluated across nine test sets to assess their causal inference capabilities, generalization (i.e., performance under distribution shift), internalization (i.e., whether the LLM truly understands the underlying causal inference theorems), and robustness to practical stressors (i.e., noise and missing data) relevant to real-world settings.

The comprehensive experiments on nine diverse testing sets using \base~demonstrate that proper post-training can enable smaller-scale LLMs to function as strong \emph{causal reasoners} that surpass the larger-scale LLM. Specifically, GRPO emerges as the most effective method, achieving an impressive 93.5\% accuracy on the CaLM benchmark \citep{chen2024causal}, whereas \ds~and \openai~reach only 57.0\% and 55.4\%, respectively. Moreover, the post-trained LLMs not only excel at causal inference but also exhibit strong generalization to distribution shift, effective internalization of knowledge, and robustness to noise.

To summarize, our main contributions are: 

\begin{enumerate}[leftmargin=*]
    \item To the best of our knowledge, this is the first work to thoroughly investigate the effects of current mainstream post-training methods on the causal inference abilities of LLMs.
    \item We introduce the \sys~dataset, comprising (i) the first training set designed to systematically enhance LLMs as \emph{causal reasoners}. This training set encompasses seven distinct tasks and is adaptable to five different post-training methods. And (ii) a suite of five test sets that evaluate LLMs along three dimensions: generalization, internalization, and robustness.
    \item We conduct comprehensive experiments to validate the causal inference capabilities, generalization, internalization, and robustness of LLMs trained using various post-training methods.

\end{enumerate}

%% file: sections/3_methodology.tex
\section{Methodology}
Our method proceeds in four main steps. First, we generate training data from synthetic SCMs (Sec. \ref{method:training_data}) and create five specialized testing sets to thoroughly evaluate the \emph{causal reasoner}'s ability (Sec. \ref{method:testing_data}). We collectively refer to the entire corpus as \sys. 
Next, we adopt a two-stage training strategy: 
cold-starting the LLM with SFT on a small amount of data (Sec. \ref{method:cold_start}), followed by five various post-training methods (i.e., SFT, PPO, GRPO, DPO and KTO) to enhance the LLM's causal inference ability (Sec. \ref{method:post_training}).

\subsection{Training Dataset Generation}
\label{method:training_data}


Our approach is two-fold: we first generate a base dataset and then make fine-grained adjustments according to the requirements of each post-training method.

\subsubsection{Base Dataset Construction}
\label{section:gen}

While the focus of \citet{chen2024causal} is the testing set, the field still lacks sufficient training data for causal inference. We address this by replicating their data construction method and using it as a foundation to build extended training sets tailored for various post-training methods.
To achieve this, we employ the following four steps to construct our base dataset:

\paragraph{Step 1: generating DAGs.}
We create the backbone structure for our SCMs by randomly generating 10-node DAGs, a graph size widely adopted in current research \citep{jin2023cladder,chen2024causal,chen-etal-2024-clear}.
%
\paragraph{Step 2: semantifying nodes.}
Following \citet{jin2023cladder} and \citet{chen2024causal}, we assign meaning to the nodes in three distinct ways to ensure diversity: 
(a) \textbf{Real:} Nodes receive semantically meaningful labels, and their causal relationships are coherent based on domain knowledge.
(b) \textbf{Random:}  Nodes receive semantically meaningful labels, but the causal relationships between them are randomized. 
(c) \textbf{Fake:} The nodes' labels are assigned as stochastic four-letter strings.

\paragraph{Step 3: determining SCMs.}
We model the underlying functions of the SCMs with single-layer perceptrons, aligning the defined causal relationships with the perceptron parameters. To be specific, the SCM function for a node $X$ can be written as:
\begin{align*}
&X := f_{X}\left({PA}_{X}^{1}, {PA}_{X}^{2}, \ldots, {PA}_X^{k}, U_{X}\right) \\
&= \left\{
\begin{array}{ll}
0,  U_{X} - g_{X}\left({PA}_{X}^{1}, \ldots, {PA}_X^{k}\right) > 0 \\
1,  \text{otherwise}
\end{array}
\right.
\end{align*}
where ${PA}_X^i$ denotes the parent nodes of X, $U_X$ is an independent random variable uniformly distributed from $[0,1]$ and $g_{X}: \{0,1\}^k \rightarrow [0,1]$ is a function to be determined. In this paper, we model it by a single-layer perceptron as follows:
\begin{align*}
    g_{X}&\left({PA}_{X}^{1}, {PA}_{X}^{2}, \ldots, {PA}_X^{k}\right)= \nonumber\\
    &\text{sigmoid}(b_X+\sum_{i=1}^kw_X^i{PA}_{X}^{i}).
\end{align*}
Both $b_X$ and $w_X^i$ are generated randomly, and we ensure the sign of $w_X^i$ is consistent with the assigned relationship between X and $PA_X^i$.

\paragraph{Step 4: generating instances.}
We approximate the required probabilities (marginal and conditional) by sampling from the SCMs. Utilizing this data and predefined templates, we generate questions, answers, and symbolic solutions for seven distinct causal tasks. 
\begin{figure}
    \begin{center}
    \includegraphics[width=\linewidth]{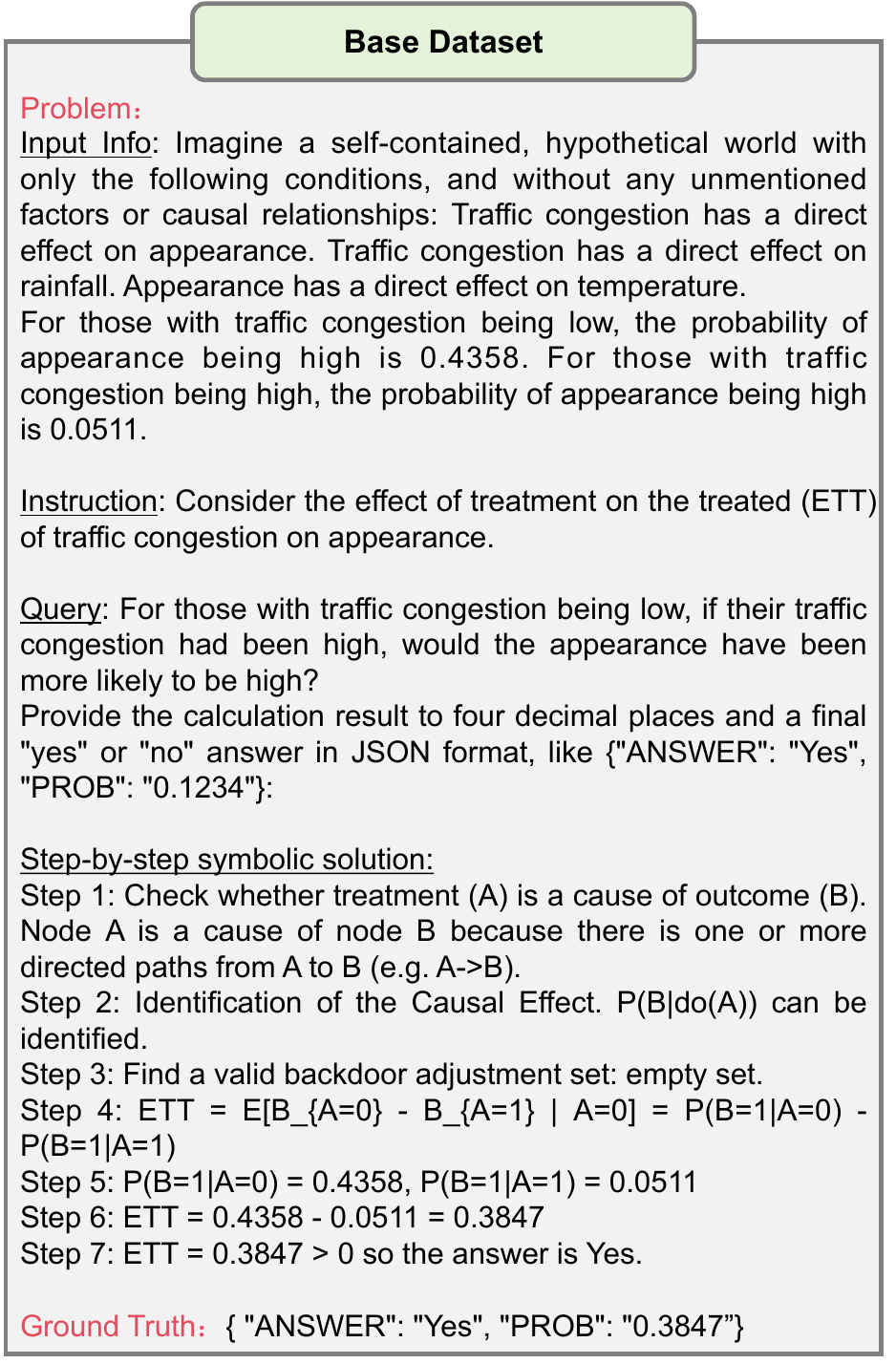}
  \end{center}
  
    \caption{An example of the base dataset.}
    \vspace{-6mm}
\label{fig:base}
\end{figure}
Figure \ref{fig:base} provides an example of the resulting data. We then leverage \texttt{DoWhy} \citep{dowhy} to identify the necessary backdoor adjustment set and mediator set, and then combine this information to formulate the symbolic solutions with several templates\footnote{To be specific, the predefined templates for questions are shown in Table \ref{tab:question_templates}.}.  At this point, the base dataset is complete and can be processed into the specific formats needed for all subsequent post-training methods.

\subsubsection{Method-Specific Adaptation}
We now proceed to describe how the base dataset is modified for various post-training methods. 

(1) \textbf{SFT.} 
For each question, we first provide a step-by-step symbolic solution to prompt the \ds~ to generate a correct and naturally phrased reasoning process and answer. The SFT dataset retains these rephrased reasoning traces and correct answers for finetuning.
(2) \textbf{Offline RL.}
On the same set of questions, we construct paired positive/negative reasoning samples.
(a) Positive: Built in the same way as SFT. 
(b) Negative: Generated by instructing the \ds~to answer questions without any step-by-step guidance. If the final answer is wrong, we use the associated reasoning as the negative sample. If the model consistently answers correctly, we select an unguided reasoning trace that is verbose, missing key steps, or misaligned with the standard solution as the negative sample.
Accordingly, DPO forms one positive–negative pair per question, while KTO collects all generated positive and negative samples and enforces a 1:1 ratio.
(3) \textbf{Online RL.}
Since GRPO and PPO rely solely on the final-answer reward and format reward (reasoning format and JSON format), chain-of-thought annotations are unnecessary. For these online RL methods, we only need to extract the questions and final answers directly from the base dataset.




\subsection{Testing Dataset Generation}
\label{method:testing_data}

To thoroughly evaluate a \emph{causal reasoner}'s ability, we create five specialized testing sets based on three distinct perspectives: 
(1) Generalization. To test whether the LLM truly understands the meaning of the questions rather than simply memorizing the specific phrasing found in the training data, we design the \reph. 
(2) Internalization. To determine if the LLM truly understands the underlying causal inference theorems rather than relying on superficial shortcuts, we create the \tf~and \hard. 
(3) Robustness. To assess the LLM's robustness under the non-ideal, noisy data conditions common in the real world, we construct the \redun~and \lack.
To build these new testing sets, we take the ATE, CDE, ETT, NDE, NIE, PN, and PS tasks from the original CaLM dataset \citep{chen2024causal} and modify them in various ways.



\textbf{\reph.} The \reph~is constructed by prompting the \ds~to rephrase the whole problem. The meaning, probabilities, and step-by-step solving procedure remain preserved. An example is shown in Figure \ref{fig:rephrased_prompt}.

\textbf{\tf~and \hard.} The \tf~paraphrases the original whole problem and preserves its semantics and probabilities, but intentionally omits the instruction part (previously shown in Figure \ref{fig:base}) that would otherwise reveal the specific task type. This design prevents models from relying on explicit task cues. The questions in the \hard~can only be solved by applying the backdoor criterion. We specifically avoid cases such as those where no causal relationship exists between the cause and effect, or where confounders are absent. This design ensures that the model cannot rely on spurious correlations to get the correct answer and must understand the underlying causal structure. The examples are shown in Figure \ref{fig:omitted_prompt} and \ref{fig:deconfounding_prompt}.

\textbf{\redun~and \lack.} 
In the real world, causal inference problems posed by non-experts are highly likely to contain either redundant or insufficient information. To test robustness against this, we construct \redun~(by adding two correct but useless conditions) and \lack~(by removing two necessary conditions). We hypothesize that a true \emph{causal reasoner} will successfully disregard the redundant data in \redun~and identify the key missing information in \lack. The examples are in shown in Figure \ref{fig:redundant_prompt} and \ref{fig:insufficient_prompt}.

\subsection{Cold Start}
\label{method:cold_start}
Before further employing post-training methods, we first conduct SFT to cold start the base model. The optimization objective is shown as follows:
$
    \mathcal{J}_{\text{SFT}}(\theta) = \mathbb{E}_{(x,y) \sim D}[\log \pi_{\theta}(y |x)],
$
where $\pi_{\theta}$ is the current model policy, $D$ is the dataset, $x$ is the input, and $y$ is the target completion.
\subsection{Post-training Methods}
\label{method:post_training}


Online RL methods update the model by using feedback on its own rollouts. Both PPO and GRPO learn a policy by estimating advantages for state–action pairs and ascending the surrogate objective to maximize expected rewards. Mathematically, they aim to maximize the following function:
\begin{align*}
   &\mathcal{J}(\theta) =\mathbb{E}_{o_t, a_t \sim \pi_{\theta_{\text{old}}}} \left[ \frac{1}{G} \sum_{i=1}^{G} \frac{1}{|o_i|} \sum_{t=1}^{|o_i|}\min  \right. \nonumber\\
   &\left. \left(r_t(\theta)\hat{A}_{i,t}, \text{clip}(r_t(\theta), 1-\epsilon_{low}, 1+\epsilon_{high})\hat{A}_{i,t}\right) \right] \nonumber\\
   &- \beta D_{KL}(\pi_{\theta} | \pi_{\text{ref}})
\end{align*}
where $\pi_{ref}$,$\pi_{\theta}$ and $\pi_{\theta_{old}}$ are reference, current and old policy, $A_{i,t}$ is an estimated advantage, $r_t(\theta)$ is the log-likelihood ratio and $\epsilon_{low},\epsilon_{high}$ are clip ratio high and low. The key distinction between PPO and GRPO lies in how the advantage is calculated: PPO uses a separate critic network to estimate the advantage, while GRPO uses the mean reward of the responses generated by the current policy.

DPO and KTO use a fixed, pre-collected dataset and train the LLM to separate good from bad behavior. Their optimization objectives are shown as follows:
\begin{align*}
&\mathcal{J}_{\text{KTO}}(\theta) = 
\mathbb{E}_{(x,y) \sim D_{\text{desirable}}} \nonumber\\
&\left[ 
\lambda_d \left(1 - \text{sigmoid}\left(\beta \log \frac{\pi_{\theta}(y|x)}{\pi_{\text{ref}}(y|x)} - z_{\text{ref}}\right)\right)
\right] \nonumber\\
&+\mathbb{E}_{(x,y) \sim D_{\text{undesirable}}} \nonumber\\
&\left[ 
\lambda_u \left(1 - \text{sigmoid}\left(z_{\text{ref}} - \beta \log \frac{\pi_{\theta}(y|x)}{\pi_{\text{ref}}(y|x)}\right)\right)
\right],
\end{align*}

\begin{align*}
&\mathcal{J}_{\text{DPO}}({\theta}) =  \mathbb{E}_{(x, y_w, y_l) \sim \mathcal{D}} \\
&\left[ \log \sigma \left( \beta \log \frac{\pi_{\theta}(y_w|x)}{\pi_{\text{ref}}(y_w|x)} - \beta \log \frac{\pi_{\theta}(y_l|x)}{\pi_{\text{ref}}(y_l|x)} \right) \right],
\end{align*}
where $z_{\text{ref}}$ is the reference point, $\lambda_d,\lambda_u,\beta$ are hyperparameters.

%% file: sections/4_experiment.tex
\section{Experiment}
\subsection{Setup} 
\textbf{Baselines.} We consider a wide range of baselines, including \llama~\citep{meta2024llama3}, \qwen~\citep{qwen3}, \base~\citep{guo2025deepseek}, \ds~\citep{guo2025deepseek}, \gemini~\citep{deepmind2025gemini}, \openai~\citep{openai2025o3}.



\textbf{Datasets.} Our evaluation covers a total of nine datasets: the five novel datasets we constructed in Sec. \ref{method:testing_data}, the lite version of CaLM \citep{chen2024causal} (focusing on numerical tasks for ATE, CDE, ETT, NDE, NIE, PN, and PS), and three external math benchmarks: MATH 500 \citep{hendrycksmath2021}, Minerva Math \citep{lewkowycz2022solving}, and AMC 2023 \citep{amc2023math}.


\textbf{Prompts.} We employ a basic CoT prompt (i.e., \texttt{<question}, \texttt{Let's think Step by Step>}) for all test sets. For \lack, we further augment the prompt to explicitly instruct the LLM: \emph{If the condition is not enough to solve the question, output `LACK\_CONDITION' as final answer}.

\textbf{Metrics.} The evaluation metric is accuracy. All questions are assessed with exact-match scoring. To ensure the reliability of the results, we conduct five independent runs and report the mean accuracy across all trials.

\textbf{Implementation details.} 
For SFT, we train 3 epochs on 3500 samples with LoRA \citep{hu2022lora}. For DPO, we train 3 epochs on 3500 preferred and dis-preferred pairs. The hyperparameter $\beta$ is set to 0.1. For KTO, we train 3 epochs on 7000 samples with preference labels. The hyperparameter $\beta$ is set to 0.1, $\lambda_d$ and $\lambda_u$ are both set to 1. For GRPO, we train for three epochs on a dataset of 3500 questions. We set $\beta$ to 0, $\epsilon_{low}$ to 0.2, and $\epsilon_{high}$ to 0.28 and use rejection sampling \citep{yu2025dapo}. For PPO, we train for three epochs on a dataset of 3500 questions. We set $\beta$ to 0.001, $\epsilon_{low}$ to 0.2, and $\epsilon_{high}$ to 0.28 and use rejection sampling.

\subsection{Main Results} 
\label{section:be}
\begin{table*}[htb]
\fontsize{9}{12}\selectfont
\centering
\begin{tabular}{lllllllll}
\toprule

\multicolumn{1}{c}{\bf LLM} &\multicolumn{1}{c}{\bf ATE} & \multicolumn{1}{c}{\bf CDE} & \multicolumn{1}{c}{\bf ETT} & \multicolumn{1}{c}{\bf NDE} &\multicolumn{1}{c}{\bf NIE} & \multicolumn{1}{c}{\bf PN} & \multicolumn{1}{c}{\bf PS}  & \multicolumn{1}{c}{\bf Avg.}
\\ 
\hline
\llama & 0.572 & 0.372 & 0.288 & 0.430 & 0.200 & 0.010 & 0.010 & 0.269 \\
\qwen & 0.004 & 0.000 & 0.180 & 0.230 & 0.000 & 0.000 & 0.000 & 0.059 \\
\ds & 0.740 & 0.540 & 0.220 & 0.460& 0.450 & 0.780& 0.800& 0.570\\
\gemini & 0.760& 0.710&0.320& 0.590& 0.470& 0.240 & 0.050 & 0.448 \\
\openai & 0.840& 0.590& 0.300& 0.430 & 0.720& 0.450 & 0.550 & 0.554\\
\hline
\base & 0.594 & 0.364 & 0.210 & 0.442 & 0.212 & 0.014 & 0.066 & 0.272 \\

Cold Start Base & 0.634 & 0.550 & 0.156 & 0.294 & 0.434 & 0.788& 0.714& 0.510 \\
\hline
SFT & 0.852 & 0.828 & 0.470 & 0.560 & 0.604 & 0.858 & 0.766 & 0.702 \\
DPO & 0.656 & 0.514 & 0.198 & 0.282 & 0.510 & 0.806 & 0.708 & 0.524 \\
KTO & 0.716 & 0.674 & 0.232 & 0.412 & 0.472 & 0.812 & 0.700 & 0.574  \\
PPO & \underline{0.972} &\underline{0.982} & \underline{0.806} & \underline{0.926} & \underline{0.924} & \textbf{0.940} & \textbf{0.902} & \underline{0.921}\\
GRPO & \textbf{0.990} & \textbf{0.994} & \textbf{0.900} & \textbf{0.940} & \textbf{0.930} & \underline{0.928} & \underline{0.866} & \textbf{0.935} \\
\bottomrule

\end{tabular}
\caption{Comparison of different post-training methods and a wide range of baselines. Best results are in \textbf{bold}, the second best results are \underline{underlined}.}
\label{tab:overall}
\end{table*}

Table \ref{tab:overall} presents a comparison of our different training approaches with baseline models. We draw the following conclusions:
(1) Through appropriate post-training, it is possible to build \emph{causal reasoners} using smaller-scale LLMs that outperform larger-scale LLMs. We exclusively train the \emph{causal reasoner} on a 14B-scale LLM (i.e., \base). Among these methods, DPO performs the least effectively, achieving a score of only 52.4\%, while the best-performing GRPO reaches an impressive 93.5\%.
Despite DPO’s lower performance compared to other methods, it still enables the \emph{causal reasoner} to perform on a par with \ds. This demonstrates that current advanced post-training methods can significantly enhance the causal inference capabilities of LLMs.
(2) On average performance, GRPO proves to be the most effective method for building \emph{causal reasoners}. After training with GRPO, the average performance of the LLM reaches 93.5\%. This is 42.5\% higher than \base, and more than 23.3\% over SFT.
%
%
(3) All post-training methods improve the LLM's causal inference performance to varying degrees. Relative to the cold-start baseline, SFT yields a 19.2\% gain, DPO 1.4\%, KTO 6.4\%, PPO 41.1\%, and GRPO 42.5\%.
(4) In general, online RL methods (GRPO, PPO) demonstrate statistically significant superiority over offline RL methods (DPO, KTO) and SFT. The offline RL post-training methods and SFT gain a 3.9\% and 19.8\% improvement over the model only with a cold start respectively, while the online methods gain a surprising 41.8\% improvement. 



\subsection{Generalization}

\begin{figure}[ht]
\centering
    \subfigure[Performance]{
  \begin{minipage}[t]{\linewidth}
    \centering
    \includegraphics[width=0.8\linewidth]{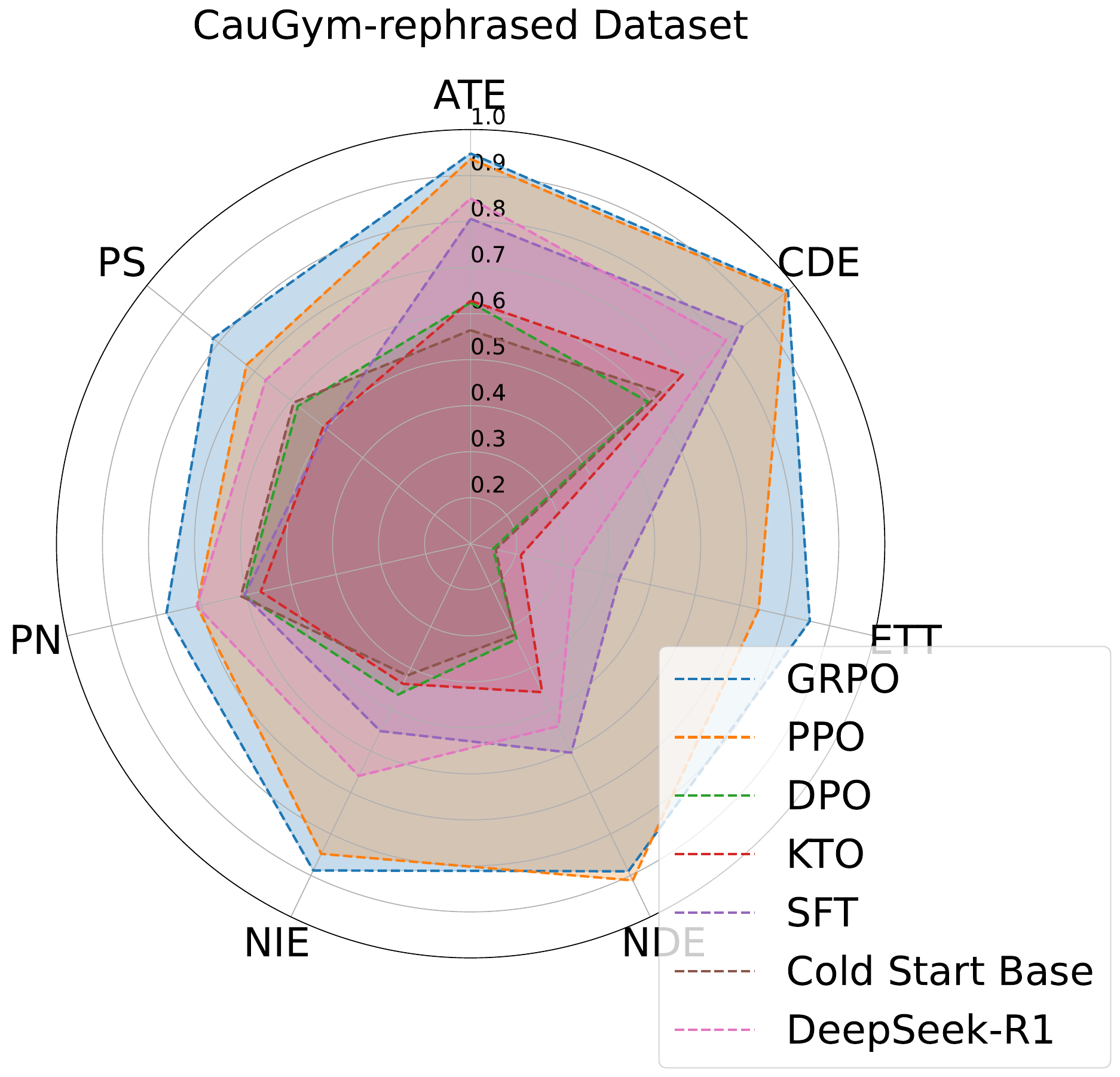}
    \label{fig:reph}
  \end{minipage}
  }
  \subfigure[Difference]{
  \begin{minipage}[t]{\linewidth}
    \centering
    \includegraphics[width=0.8\linewidth]{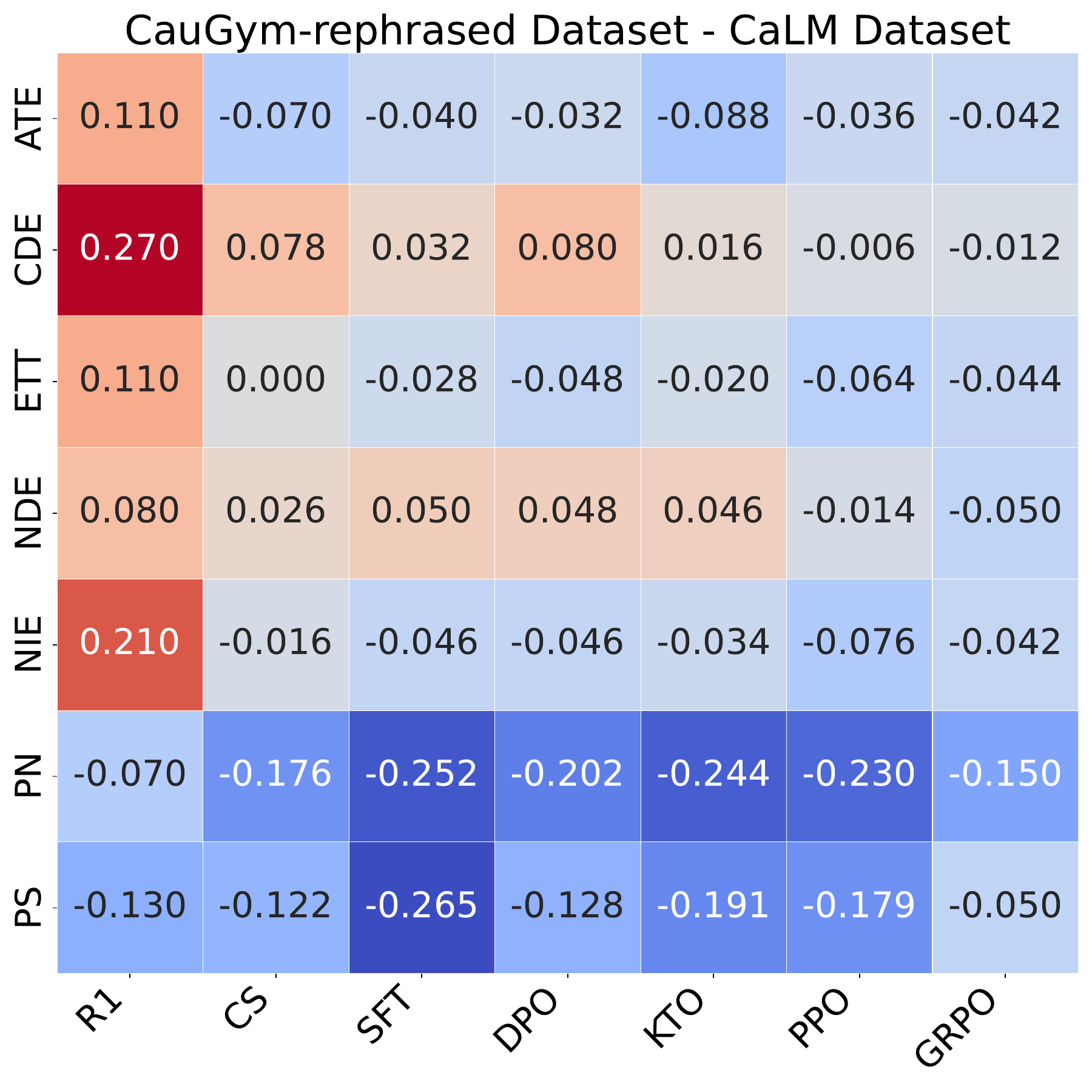}
    \label{fig:reph_diff}
  \end{minipage}
  }
  \caption{(a) Model performance on \reph. (b) Model performance difference between the dataset and CaLM. ``R1'' denotes \ds, ``CS'' denotes Cold Start Base.}
\end{figure}

Figure \ref{fig:reph} and Figure \ref{fig:reph_diff} present the performance of our different training approaches on the \reph~and show the performance difference between these approaches on the CaLM dataset and the \reph. We draw the following conclusions:
(1) Online RL methods still maintain their superiority on \reph. The average performance of online RL methods GRPO and PPO still reaches 85.7\%, while the average performance of offline RL methods and SFT reaches only 48.9\% and 62.2\%.
(2) Post-training methods are robust to paraphrasing. The average performance drop of post-training methods after paraphrasing is only 6.8\%, while that of the model only with a cold start is 4.0\%. 
(3) \ds~is also robust to paraphrasing. The average of its performance is 65.2\%, which is even 8.2\% higher than its performance on CaLM.

\begin{table}[h]

\centering
\resizebox{0.95\columnwidth}{!}{\begin{tabular}{ccccc}
\toprule
\multicolumn{1}{c}{\bf LLM} &\multicolumn{1}{c}{\bf MATH 500} &\multicolumn{1}{c}{\bf AMC 2023}& \multicolumn{1}{c}{\bf Minerva Math} & \multicolumn{1}{c}{\bf Avg.}\\
\hline
DS & 0.938 & 0.867 &0.375 & 0.727\\
SFT & 0.938 & 0.892 &0.397 & 0.742 \\
DPO & 0.926 & 0.843 &0.378 & 0.715\\
KTO & 0.932 & 0.880 &0.368 & 0.726\\
PPO & 0.924 & 0.855 &0.404 & 0.727\\
GRPO& 0.924 & 0.855 &0.408 & 0.729\\
\bottomrule
\end{tabular}
}
\caption{Performance on three math datasets, where DS stands for \base.}
\label{tab:math}
\end{table}

Table \ref{tab:math} represents the performance of post-training methods on the math test datasets. The performances of all methods are close to the performance of \base, whose maximum difference is less than 2.0\%. The result shows that employing post-training on causal inference does not degrade LLM math ability.

\subsection{Internalization}
\label{section:internalization}

\begin{figure}[ht]
\centering
    \subfigure[Performance]{
  \begin{minipage}[t]{\linewidth}
    \centering
    \includegraphics[width=0.8\linewidth]{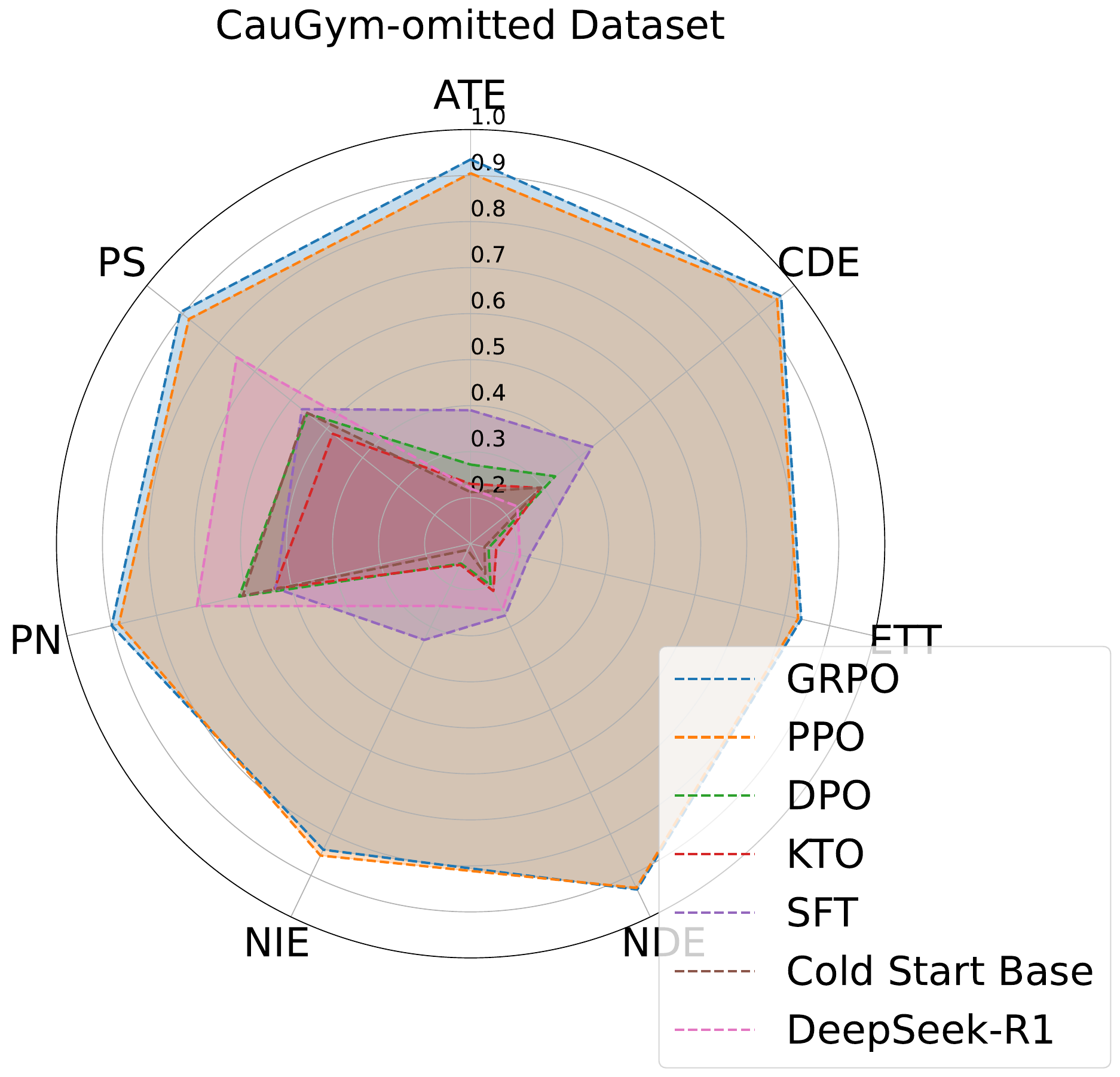}
    \label{fig:tf}
  \end{minipage}
  }
  \subfigure[Difference]{
  \begin{minipage}[t]{\linewidth}
    \centering
    \includegraphics[width=0.8\linewidth]{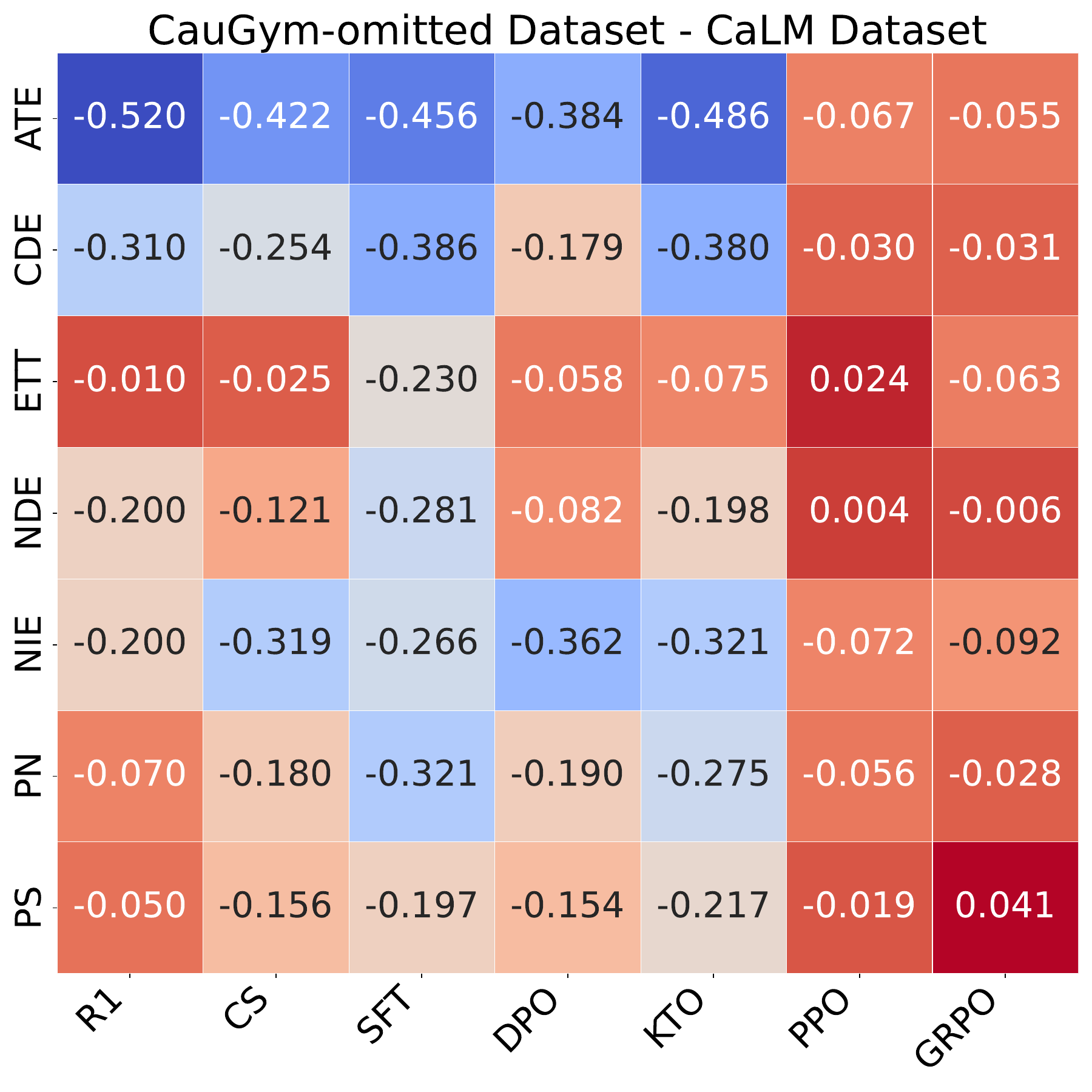}
    \label{fig:tf_diff}
  \end{minipage}
  }
  \caption{(a) Model performance on \tf. (b) Model performance difference between the dataset and CaLM. ``R1'' denotes \ds, ``CS'' denotes Cold Start Base.}
\end{figure}

Figure \ref{fig:tf} and Figure \ref{fig:tf_diff} present the performance of our different training approaches and \ds~on the \tf~and show the performance difference between these approaches on the CaLM dataset and the \tf. We find that:
(1) Online RL methods still maintain their superiority on \tf. The average performance of online RL methods still reaches 89.6\%, while the average performance of offline RL methods and SFT reaches only  30.9\% and 39.5\%.
(2) Online RL methods are robust to removing instructions, but offline RL methods and SFT are not. The average performance drop of online RL after omitting is only 3.2\%, while the average performance drop of offline RL and SFT reach 24.0\% and 30.7\% respectively. In contrast, the average performance drop of cold start base model is  17.1\%.
(3) \ds~is not robust to removing instructions. The average of its performance is 37.5\%, which is 19.5\% lower than its performance on CaLM.

\begin{figure}[h]
    \centering
    \includegraphics[width=
    \linewidth]{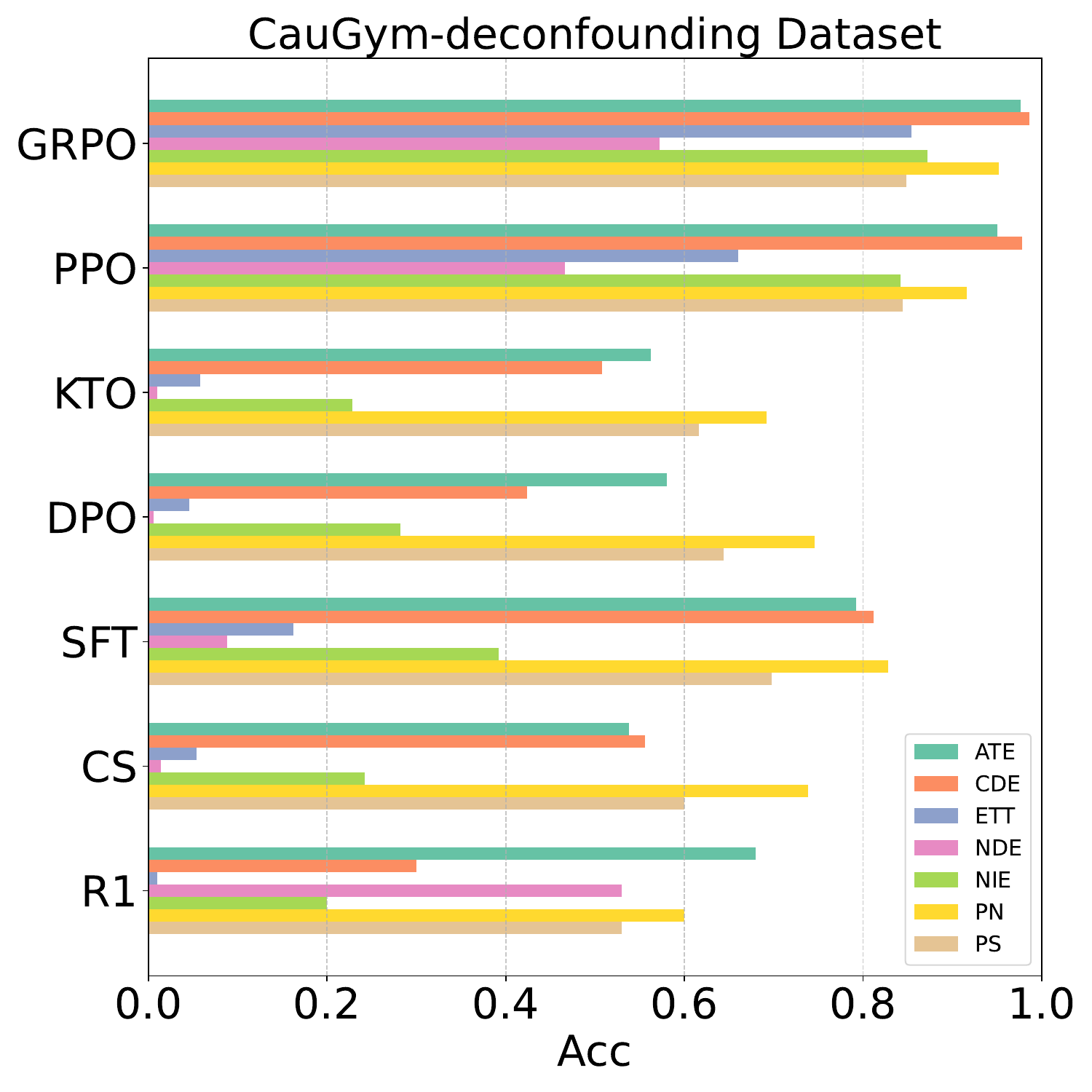}
    \caption{Model performance on \hard, ``R1'' denotes \ds, ``CS'' denotes Cold Start Base.}
    \label{fig:hardest}
\end{figure}

Figure \ref{fig:hardest} presents the performance of different training approaches and \ds~on the \hard. We can conclude that:
(1) The online RL method GRPO still has the best performance. Its average performance on the \hard~reaches 86.5\%. In contrast, the average performance of KTO, DPO, SFT, PPO is 38.2\%, 39.0 \%, 53.9\% and 80.8\%. Moreover, the performance of  GRPO is higher than any other post-training method on every causal inference task.
(2) Online RL methods perform well on this dataset, while offline RL methods struggle. The average performance of cold start base model is 39.2\%. The average improvement of online RL method is 44.4\% and that of SFT is 14.7\%. However, offline RL methods show no improvement.
(3) \ds~is poor at applying the backdoor criterion. The average performance of \ds~is 40.7\%, similar to cold start base model. 

In general, the result shows that online RL methods, especially GRPO, enable LLM to understand the underlying causal structure of given questions and apply causal inference theorems independently without additional cues. It also reveals that \ds~actually struggles at identifying spurious correlations and understanding causal inference tasks.

\subsection{Robustness}

\begin{figure}
\centering
    \subfigure[\redun~dataset]{
  \begin{minipage}[t]{\linewidth}
    \centering
    \includegraphics[width=\linewidth]{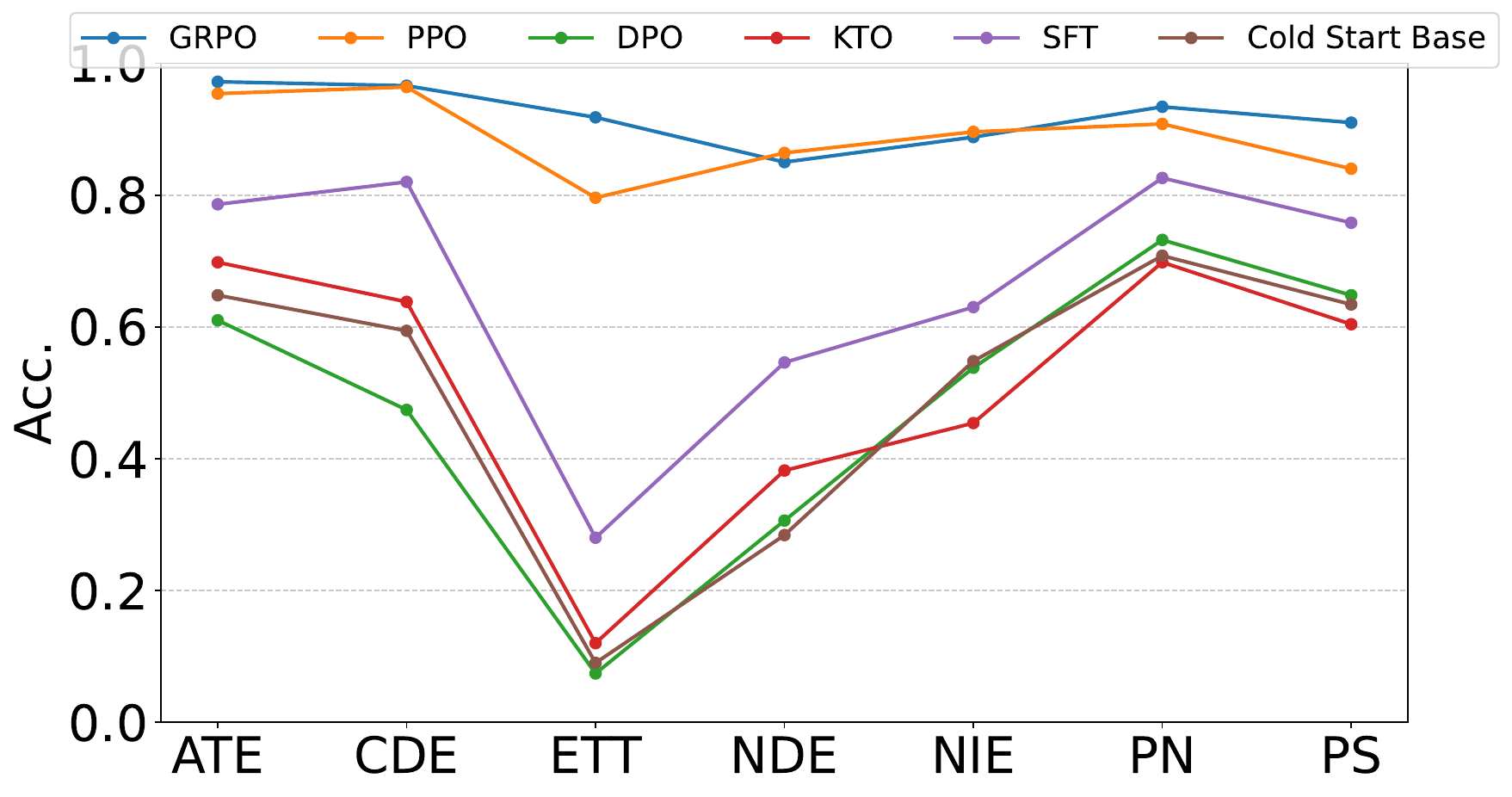}
    \label{fig:redundant}
  \end{minipage}
  }
  \subfigure[\lack~dataset]{
  \begin{minipage}[t]{\linewidth}
    \centering
    \includegraphics[width=\linewidth]{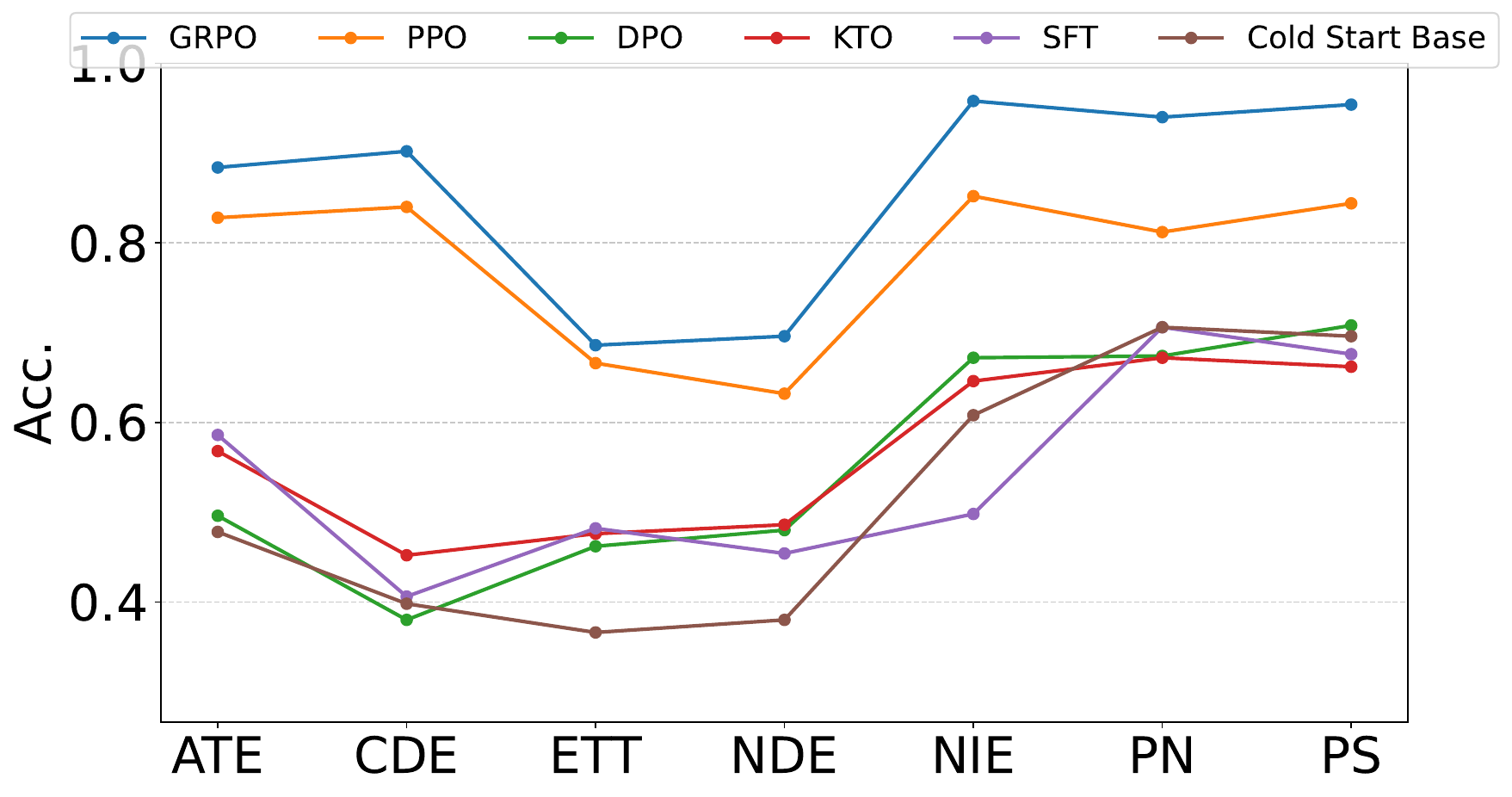}
    \label{fig:lack}
  \end{minipage}
  }
  \caption{Model performance on \redun~dataset and \lack~dataset.}
\end{figure}

Figure \ref{fig:redundant} presents the performance of different training approaches on the \redun~Dataset. We draw the following conclusions:
(1) The online RL method GRPO still performs best, which average performance on the \redun~Dataset reaches 92.0\%. In contrast, the average performance of KTO, DPO, SFT, PPO is 51.3\%, 48.3 \%, 66.3\% and 88.9\%. 
(2) Online RL methods perform well; SFT has some effect; Offline RL methods yield marginal improvements. The average performance of the model with only a cold start is 50.1\%. The average improvement of online RL methods is 40.3\% and that of SFT is 16.2\%, while there is no improvement for offline RL methods.

Figure \ref{fig:lack} presents the performance of different training approaches on the \lack~Dataset. Our key findings are as follows:
(1) The online RL method GRPO still has the best performance. Its average performance on the \lack~reaches 86.5\%. In contrast, the average performance of KTO, DPO, SFT, PPO is 56.6\%, 55.3\%, 54.4\% and 78.2\%. Moreover, the performance of  GRPO is higher than any other post-training method on every causal inference task.
(2) Online RL methods perform well on this dataset, while SFT and offline RL methods struggle. The average performance of the model with only a cold start is 51.9\%. The average improvement of online RL method is 30.4\%, while that of SFT and offline RL methods is about 4.7\% and 2.9\%, which are negligible.
In general, the result shows that online RL methods significantly improve a LLM's ability to identify the appropriate data to solve a given question. Offline RL and SFT, however, enhance this ability less.

\subsection{Ablation}
\subsubsection{Base Model Ablation}
To further demonstrate the advantage of online RL methods in improving the causal inference capability of LLMs, we further train and evaluate multiple base models, including Mistral-7B and DeepSeek-R1-Distill-Llama-8B. We apply GRPO, DPO, and SFT to these two models, and the result is shown in Table \ref{tab:model_comparison}. The strong performance of GRPO models further confirms that its effects consistently extend beyond a single model.
\begin{table}[!ht]
    \small
    \centering
    \begin{tabular}{cccccc}
        \hline
        \textbf{Method} &\textbf{CaLM} & \textbf{Omitted} & \textbf{Red} & \textbf{Reph} \\
        \hline
        MS-Base& 0.151 & 0.117 & 0.124 & 0.163 \\
    
        MS-DPO & 0.067 & 0.038 & 0.048 & 0.048 \\
   
        MS-SFT & 0.411 & 0.266 & \textbf{0.421} & 0.374 \\
    
        MS-GRPO & \textbf{0.511} & \textbf{0.479} & 0.363 & \textbf{0.479} \\
    \hline
        LMA-Base & 0.213 & 0.152 & 0.136 & 0.256 \\
  
        LMA-DPO & 0.357 & 0.260 & 0.311 & 0.319 \\
 
        LMA-SFT & 0.560 & 0.315 & 0.525 & 0.490 \\
  
        LMA-GRPO & \textbf{0.861} & \textbf{0.843} & \textbf{0.829} & \textbf{0.811} \\
        \hline
    \end{tabular}
    \caption{Performance comparison across different base LLMs and post-training methods, where MS-7B stands for Mistral-7B, LMA stands for DeepSeek-R1-Distill-Llama-8B, Red stands for \sys-redundant and Reph stands for \sys-rephrased.}
    \label{tab:model_comparison}
\end{table}
\subsubsection{Model Scale Ablation}
To further investigate whether online RL maintains its advantage across varying model scales, we apply DPO, SFT, and GRPO algorithms to both DeepSeek-R1-Distill-Qwen-7B and DeepSeek-R1-Distill-Qwen-32B models. In table \ref{tab:scale_comparison}, we evaluate the fine-tuned models alongside their base counterparts across four benchmarks: CaLM, \tf, \redun, and \reph. The results indicate that online RL consistently achieves superior performance in both 7B and 32B scales, further demonstrating the generalizability of RL in enhancing causal inference. Notably, the performance of the 14B and 32B models is close after fine-tuning across all four datasets. This suggests that for improving causal reasoning, selecting an appropriate algorithm may be more critical than simply increasing model scale.
\begin{table}[t]
    \centering
    \fontsize{9}{12}\selectfont
    \begin{tabular}{cccccc}
        \hline
        \textbf{Method} &\textbf{CaLM} & \textbf{Omitted} & \textbf{Red} & \textbf{Reph} \\
        \hline
        7B-Base& 0.154 & 0.103 & 0.100  & 0.234\\
    
        7B-DPO &0.373 & 0.189 & 0.351 & 0.295 \\
   
        7B-SFT & 0.493  &  0.274 &  0.458 & 0.390\\
    
        7B-GRPO &\textbf{0.585}& \textbf{0.509} & \textbf{0.487} & \textbf{0.501} \\
    \hline
        32B-Base & 0.296 & 0.134 & 0.561 & 0.359  \\
  
        32B-DPO & 0.583 & 0.425  & 0.545 & 0.567 \\
 
        32B-SFT & 0.727  & 0.509  &  0.687 & 0.705 \\
  
        32B-GRPO & \textbf{0.916} & \textbf{0.871} & \textbf{0.873} & \textbf{0.835} \\
        \hline
    \end{tabular}
    \caption{Performance comparison across different model scale, where 7B stands for DeepSeek-R1-Distill-Qwen-7B, 32B stands for DeepSeek-R1-Distill-Qwen-32B, Red stands for \sys-redundant and Reph stands for \sys-rephrased.}
    \label{tab:scale_comparison}
\end{table}

%% file: sections/5_related_works.tex
\section{Related Works}
\textbf{Post-training methods for LLMs.}
Due to impressive general abilities of pre-trained LLMs, recent research focuses on post‑training methods to further refine their problem‑solving skills. \citet{ouyang2022training} introduce SFT and PPO to align LLMs with human intent, marking the foundation of instruction‑following models such as InstructGPT. Building on this line of work, \citet{rafailov2023direct} and \citet{ethayarajh2024kto} propose DPO and KTO respectively, to more effectively align LLMs with human preferences without requiring explicit reward modeling. \citet{shao2024deepseekmath} introduce GRPO, an online RL method, and demonstrates remarkable success in domains such as mathematical reasoning and code generation.

\textbf{LLM causal inference ability.} 
With the emergence of LLM, researchers are now exploring how well LLMs perform at causal inference and understand causal concepts. Some research investigates the extent of LLMs' understanding of causality. \citet{zevcevic2023causal} claim that LLMs struggle to memorize and reproduce correlations of causal facts, while \citet{jin2024can} find that LLMs have the difficulty in determining causal relations from correlation statements. Other works study LLMs' performance on causal tasks. \citet{gao2023chatgpt} reveal that ChatGPT has serious hallucinations on causal reasoning, while \citet{chen2024causal} present that counterfactual problems, like ETT, NIE, PN, and their numerical causal effect estimation are still a challenge to LLMs. 


%% file: sections/6_conclusion.tex
\section{Conclusion}
Through comprehensive experiments using the novel \sys dataset, we demonstrate that post-training can transform a smaller 14B-scale LLM into a highly effective causal reasoner, outperforming larger models. Our research shows that while offline training methods equip LLMs with fundamental causal concepts, online RL methods are crucial for teaching them to apply these rules to solve complex problems. Among the methods tested, GRPO emerges as the most effective, achieving an impressive 93.5\% on the CaLM benchmark. This work establishes that online RL, and particularly GRPO, enables LLMs to generalize to rephrased questions, internalize causal theorems without explicit instructions, and robustly handle noisy data, thereby making sophisticated causal inference accessible to a broader audience.

%% file: sections/9_ethical.tex
\section{Ethical Considerations}
This research investigates the enhancement of causal inference capabilities in LLMs through targeted post-training methodologies. Our work focuses on the technical development of the \sys dataset and the evaluation of five distinct post-training approaches—SFT, DPO, KTO, PPO, and GRPO—to foster a deeper understanding of causal concepts like the backdoor criterion and ETT.

The construction of the \sys dataset relied on synthetic SCMs and randomly generated DAGs. This synthetic approach ensures that the data is free from personal identifiers, sensitive human information, or proprietary real-world data, thus upholding strict privacy protection standards. Furthermore, to mitigate potential biases, we employed three distinct node-labeling strategies: real-world semantic labels, randomized relationships, and stochastic ``fake'' strings. This diversity prevents the models from relying on superficial shortcuts or cultural biases inherent in natural language.

%% file: sections/8_limitation.tex
\section{Limitations}

\sys primarily focuses on the question: ``Can LLMs become effective causal reasoners through post-training?'' Accordingly, the objective of this work is not to optimize performance across the full spectrum of causal inference tasks, but to examine how post-training affects the model's ability to understand and apply principles such as the backdoor criterion and ETT. In this sense, the resulting models are not intended to serve as general-purpose causal inference systems. Rather, they are designed to assess the extent to which post-training alone can move LLMs toward principled causal reasoning.

%% file: sections/10_acknowledgments.tex
\section*{Acknowledgments}
This work is supported in part by the Shanghai Artificial Intelligence Laboratory and in part by the National Natural Science Foundation of China under Grant 625B2131.

%% file: sections/7_appendix.tex
\section{The Use of Large Language Models}
We use a general-purpose LLM in a limited, editorial capacity: to proofread grammar and style, help rephrase a few sentences, and suggest keywords for literature searches. All ideas, analyses, experiments, and writing decisions are our own; the LLM does not generate novel content or influence the study’s methodology or results.

\input{sections/2_preliminary}

\section{Dataset Generation Template}
Table \ref{tab:question_templates} lists all templates used for generating questions of different causal tasks.
\begin{table*}[!th]
\fontsize{9}{12}\selectfont
    \centering
    {
\begin{tabularx}{\textwidth}{|l|X|}
    \hline
    
    \textbf{Causal Tasks} & \textbf{Template} \\
    \hline
    ATE & If $\{\{\text{treatment}\}\}$ is changed to be $\{\{\text{treatment\_value}\}\}$, will the $\{\{\text{outcome}\}\}$ be more likely to be $\{\{\text{outcome\_value}\}\}$? \\
    \hline
    ETT & For those with $\{\{\text{treatment}\}\}$ being $\{\{\text{treatment\_value}\}\}$, if their $\{\{\text{treatment}\}\}$ had been $\{\{\text{not\_treatment\_value}\}\}$, would $\{\{\text{outcome}\}\}$ have been more likely to be $\{\{\text{outcome\_value}\}\}$? \\
    \hline
    CDE & Conditioned on $\{\{\text{mediator\_1}\}\}$ being $\{\{\text{mediator\_1\_value}\}\}$, $\{\{\text{mediator\_2}\}\}$ being $\{\{\text{mediator\_2\_value}\}\}$, $\ldots$, $\{\{\text{mediator\_n}\}\}$ being $\{\{\text{mediator\_n\_value}\}\}$, if $\{\{\text{treatment}\}\}$ had been $\{\{\text{treatment\_value}\}\}$, would $\{\{\text{outcome}\}\}$ have been more likely to be $\{\{\text{outcome\_value}\}\}$? \\
    \hline
    NIE & Suppose $\{\{\text{treatment}\}\}$ is held constant and the mediator changes to whatever value it would have attained under $\{\{\text{treatment}\}\}$ changing to be $\{\{\text{treatment\_value}\}\}$, would the $\{\{\text{outcome}\}\}$ have been more likely to be $\{\{\text{outcome\_value}\}\}$? \\
    \hline
    NDE & Suppose the mediator keeps constant when $\{\{\text{treatment}\}\}$ is changed to be $\{\{\text{treatment\_value}\}\}$, would the $\{\{\text{outcome}\}\}$ have been more likely to be $\{\{\text{outcome\_value}\}\}$? \\
    \hline
    PS & Given that $\{\{\text{treatment}\}\}$ was $\{\{\text{treatment\_negative}\}\}$ and $\{\{\text{outcome}\}\}$ was $\{\{\text{outcome\_negative}\}\}$, what is the lower bound and upper bound of the probability that $\{\{\text{outcome}\}\}$ would have been $\{\{\text{outcome\_positive}\}\}$ if the $\{\{\text{treatment}\}\}$ had been $\{\{\text{treatment\_positive}\}\}$? \\
    \hline
    PN & Given that $\{\{\text{treatment}\}\}$ was $\{\{\text{treatment\_positive}\}\}$ and $\{\{\text{outcome}\}\}$ was $\{\{\text{outcome\_positive}\}\}$, what is the lower bound and upper bound of the probability that $\{\{\text{outcome}\}\}$ would have been $\{\{\text{outcome\_negative}\}\}$ if the $\{\{\text{treatment}\}\}$ had been $\{\{\text{treatment\_negative}\}\}$? \\
    \hline

\end{tabularx}
}
    \caption{Question templates for different causal tasks.}\label{tab:question_templates}
\end{table*}

\section{Testing Dataset Examples}
\begin{figure*}
    \centering
    \includegraphics[width=0.9\linewidth]{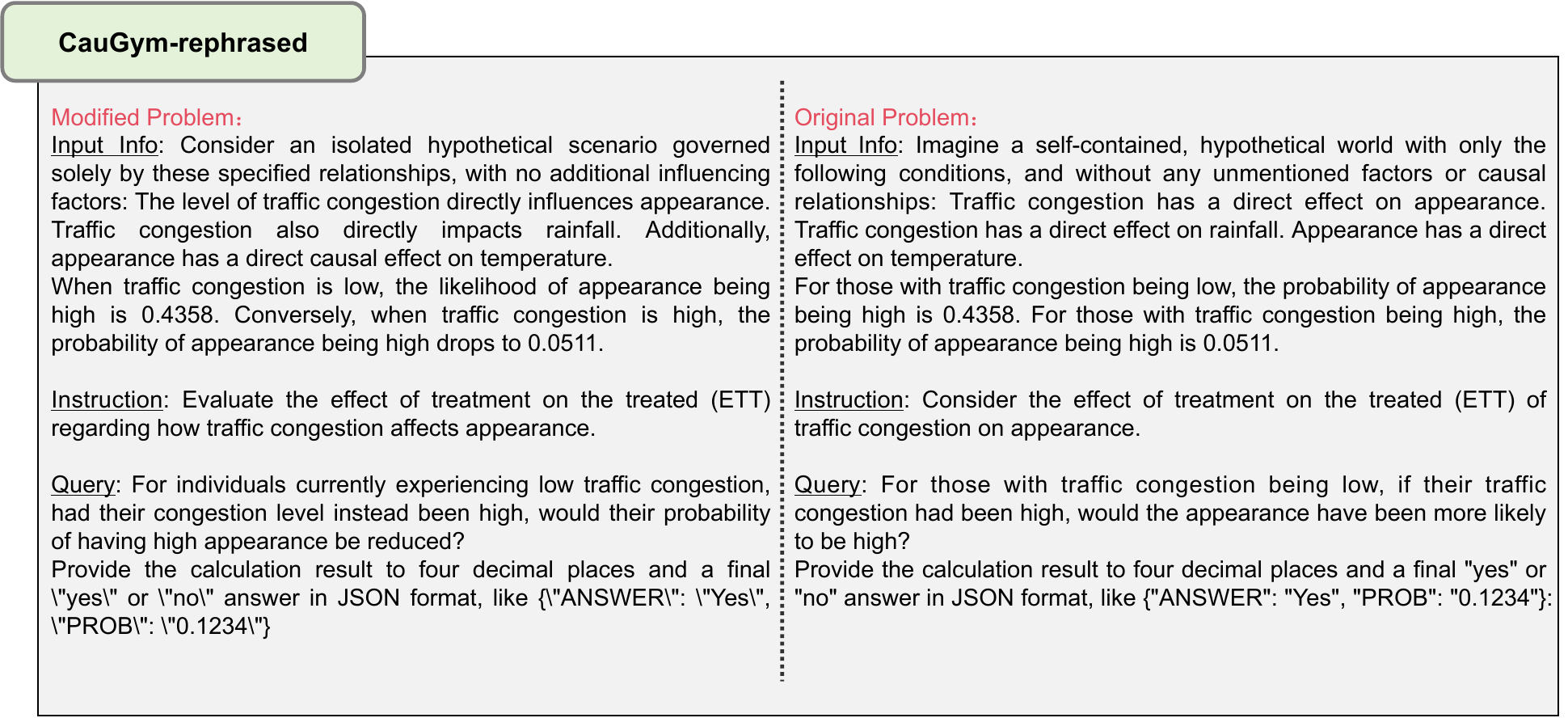}
    \caption{An example for \reph. The input info, instruction, and query are rephrased to test the robustness and overfitting tendency of the LLMs.}
    \label{fig:rephrased_prompt}
\end{figure*}

\begin{figure*}
    \centering
    \includegraphics[width=0.9\linewidth]{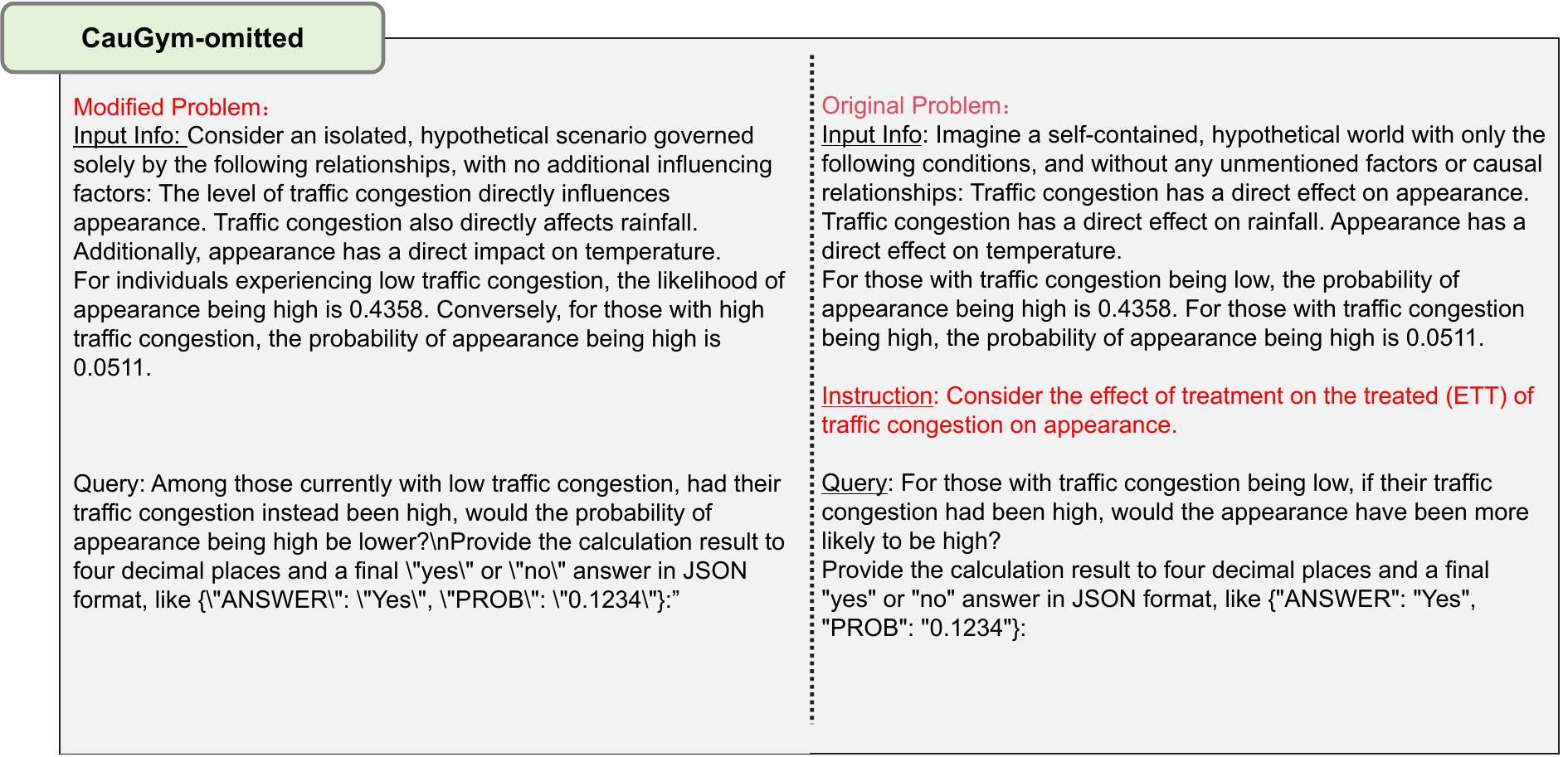}
    \caption{An example for \tf. The input info and query are rephrased while the instruction is omitted, in order to assess whether the LLMs can correctly recognize the underlying causal tasks in the problem.}
    \label{fig:omitted_prompt}
\end{figure*}

\begin{figure*}
    \centering
    \includegraphics[width=0.9\linewidth]{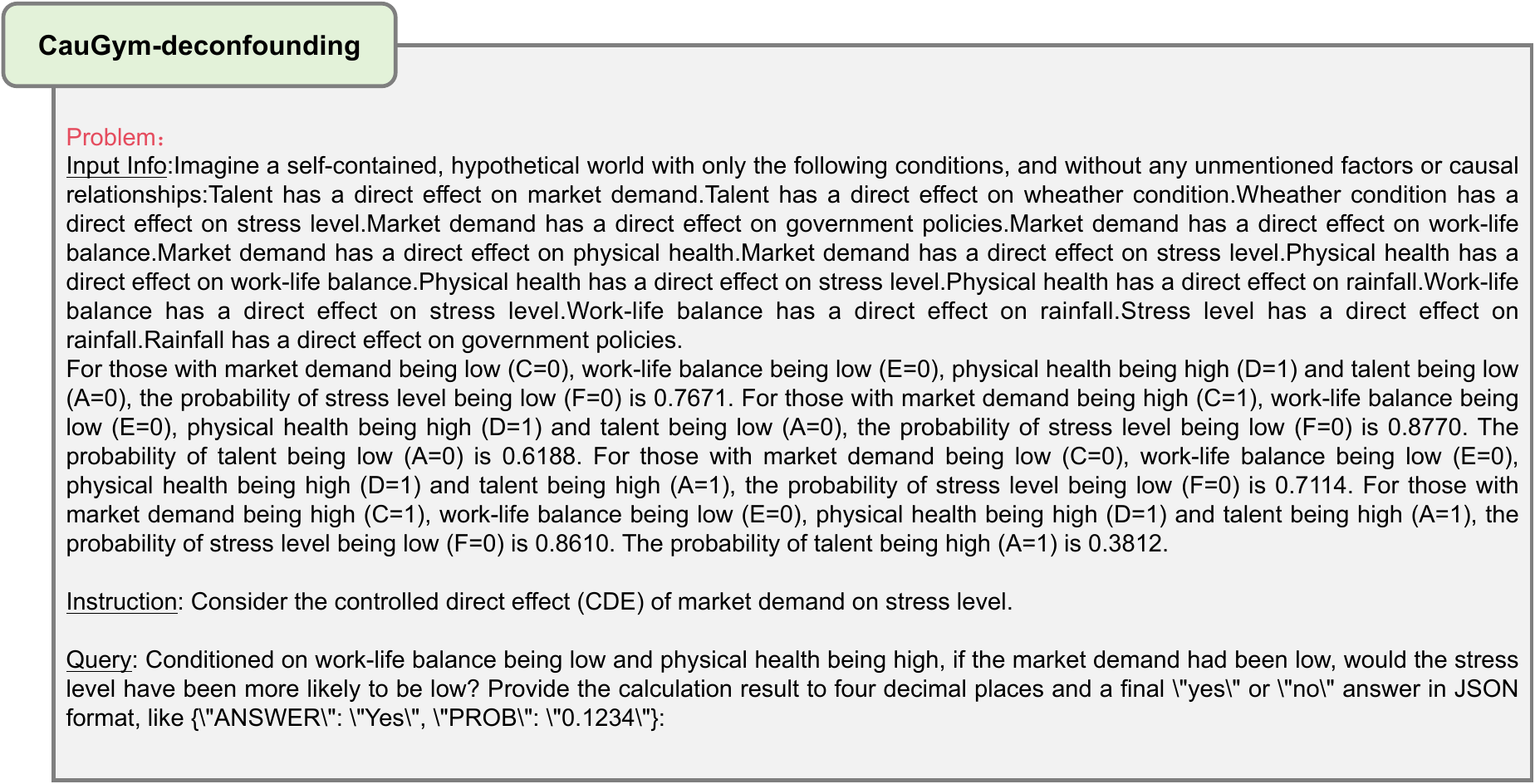}
    \caption{An example for \hard. ``Talent'' functions as a confounder in this problem. To remove its spurious influence, the backdoor criterion needs to be applied.}
    \label{fig:deconfounding_prompt}
\end{figure*}

\begin{figure*}
    \centering
    \includegraphics[width=0.9\linewidth]{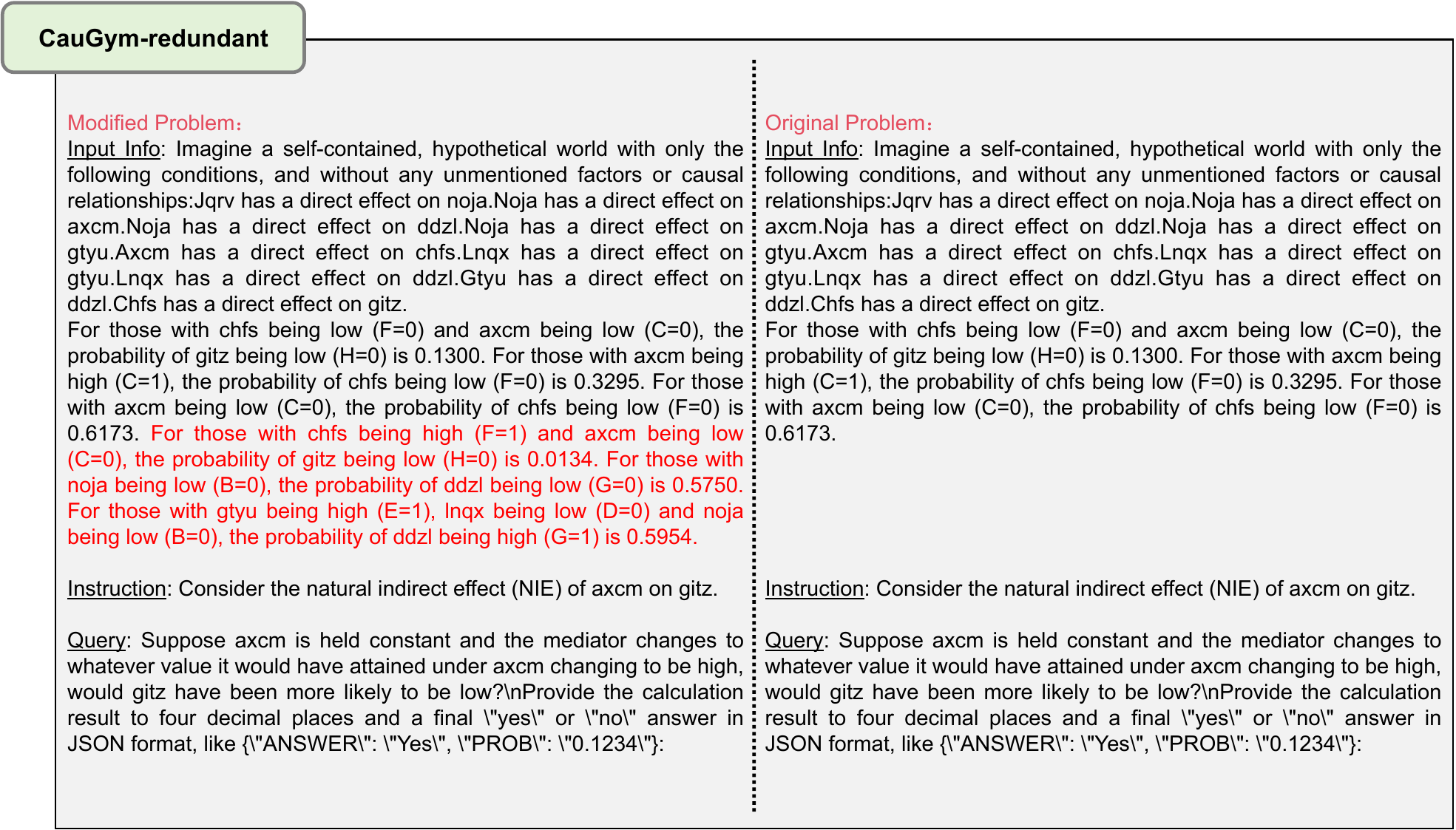}
    \caption{An example for \redun. Irrelevant conditions are introduced into the input info to test whether the LLMs remain robust against irrelevant interference.}
    \label{fig:redundant_prompt}
\end{figure*}

\begin{figure*}
    \centering
    \includegraphics[width=0.9\linewidth]{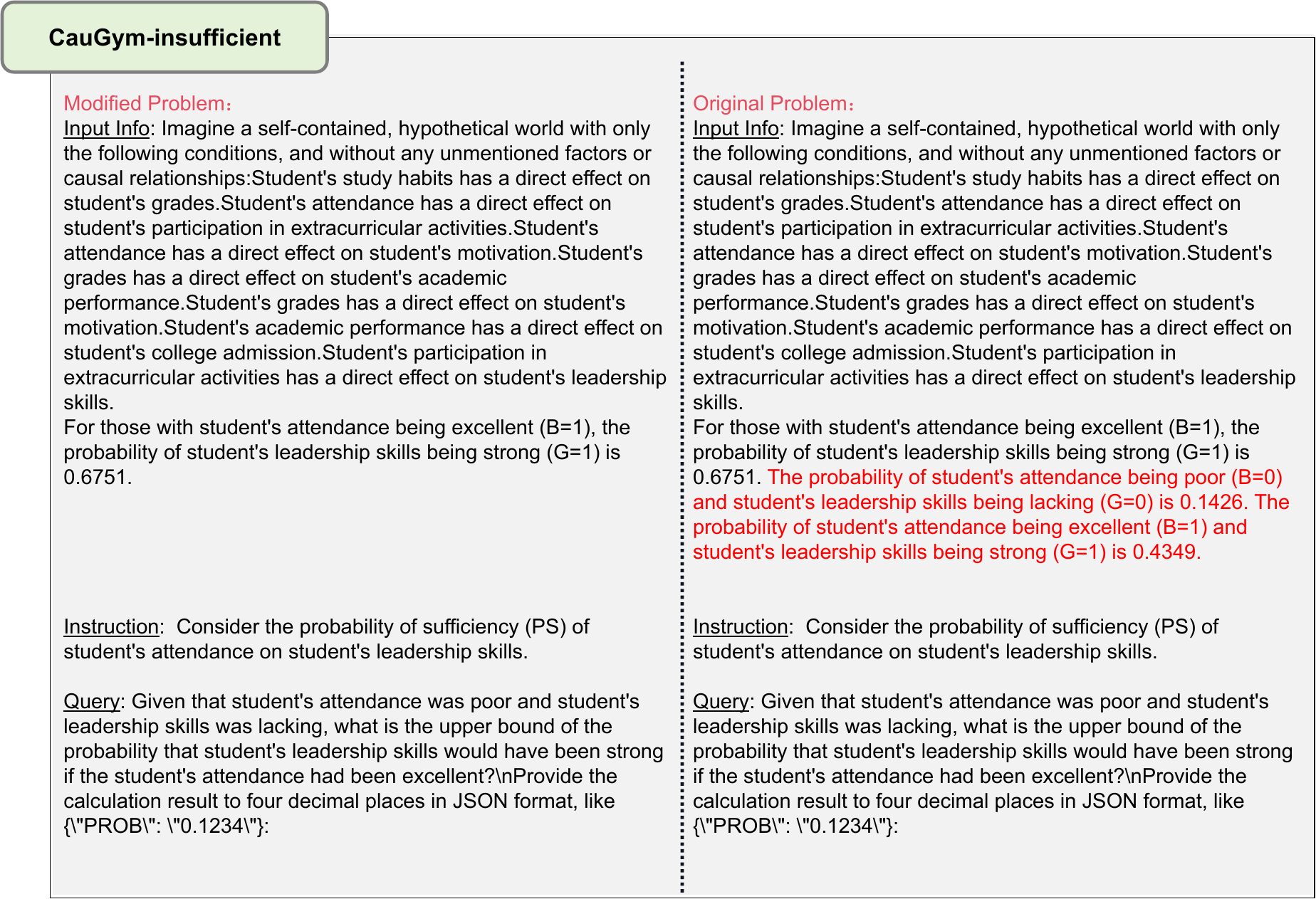}
    \caption{An example for \lack. Necessary conditional probabilities are omitted from the input info to assess whether the LLMs can detect the missing information required for correct causal inference.}
    \label{fig:insufficient_prompt}
\end{figure*}

Figure \ref{fig:rephrased_prompt}, \ref{fig:omitted_prompt}, \ref{fig:deconfounding_prompt}, \ref{fig:redundant_prompt} and \ref{fig:insufficient_prompt} provide examples of \sys~test sets. They exhibit features of each dataset.

\section{Other Model Performance on \sys}
\subsection{Base Model Variants}
\subsection{GRPO Variants}
Table \ref{tab:grpo_variants} shows the performance of two GRPO variants on \sys test sets. GRPO‑no‑think is a GRPO model trained without the reasoning format reward, while all other training configurations remain identical to the GRPO model described in the main text. Its performance exhibits only a marginal decline relative to the original GRPO model, indicating that the strong overall results primarily stem from the GRPO training framework itself rather than the particular reward design. Realistic-GRPO is trained with the same training procedure described in the main text, but the training data include only those questions in which the variable names are semantically meaningful. The significant decrease in performance suggests that the introduction of random symbolic variables induces beneficial variability, mitigating reliance on superficial semantic cues and thereby facilitating improved robustness, internalization and generalization.
\begin{table*}[!ht]
    \centering
    \fontsize{9}{12}\selectfont
    \begin{tabular}{cccccccc}
        \hline
        \textbf{Model} & \textbf{CaLM} & \textbf{Deconfounding} & \textbf{Insufficient} & \textbf{Omitted} & \textbf{Redundant} & \textbf{Rephrased} \\
        \hline
        GRPO-no-think & 0.941 & 0.834 & 0.891 & 0.651 & 0.905 & 0.868 \\
     
        Realistic-GRPO & 0.891 & 0.763 & 0.887 & 0.527 & 0.879 & 0.760 \\
        \hline
    \end{tabular}
    \caption{Performance comparison of GRPO variants on CaLM and \sys test sets.}
    \label{tab:grpo_variants}
\end{table*}

\section{Failure Analysis}
Table \ref{tab:grpo_failure}, \ref{tab:sft_failure}, and \ref{tab:cold_start_failure} illustrate the distribution and proportions of failure categories for the GRPO, SFT, and cold-start models across the CaLM, \tf, and \redun~test sets. We categorized the failures into the following types: Task Identification (Task-ID, failure in recognizing the causal task type), SCM Identification (SCM-ID, failure in identifying backdoor adjustment sets), Formula Identification (Formula-ID, failure in deriving symbolic logic), Formula Application (Formula-App, failure in executing substitutions), and Numerical Computation (Num-Comp, failure in arithmetic execution). Additionally, the category Unexpected is used to describe anomalous  behaviors, such as model self-repetition or irrelevant explanations of ETT definitions.

Our result reveals distinct failure patterns across different fine-tuned models: The cold start model primarily suffers from ``Unexpected'' failures. For SFT models, the dominant failures on test sets similar to the training dataset are Task Identification and SCM Identification; however, on more complex sets like \redun, Unexpected failures become the primary issue. In contrast, for the GRPO model, the failures are consistently centered on Task Identification and SCM Identification. Furthermore, the causal task type significantly influences the failure distribution: for PN and PS, the main bottleneck is Task Identification, whereas for other task types, SCM Identification remains the primary source of failure.

Moreover, the result also shows that GRPO outperforms other methods by significantly reducing failures in SCM identification and unexpected behaviors. This marked reduction strongly suggests an intrinsic improvement in the LLM's functional understanding of SCMs, rather than mere pattern mimicry. Furthermore, the shift from stochastic, ``unexpected'' outputs (such as repetition) to more structured, localized errors indicates that the model's reasoning patterns have become more effective and logically disciplined through the online RL process.
\begin{table*}[!ht]
\small
\begin{tabular}{llccccccc}
\toprule
\textbf{Test set} & \textbf{Type} & \textbf{Correct} & \textbf{Task-ID} & \textbf{SCM-ID} & \textbf{Formula-ID} & \textbf{Formula-App} & \textbf{Num-Comp} & \textbf{Unexpected} \\
\midrule
\multirow{7}{*}{\textbf{CaLM}} 
& ATE & 0.98 & \textbf{0.01} & \textbf{0.01} & 0.00 & 0.00 & 0.00 & 0.00 \\
& CDE & 0.99 & 0.00 & \textbf{0.01} & 0.00 & 0.00 & 0.00 & 0.00 \\
& ETT & 0.91 & 0.03 & \textbf{0.04} & 0.00 & 0.02 & 0.00 & 0.00 \\
& NDE & 0.91 & 0.00 & \textbf{0.08} & 0.01 & 0.00 & 0.00 & 0.00 \\
& NIE & 0.93 & 0.00 & \textbf{0.05} & 0.00 & 0.02 & 0.00 & 0.00 \\
& PN  & 0.94 & 0.00 & 0.01 & 0.00 & \textbf{0.05} & 0.00 & 0.00 \\
& PS  & 0.87 & 0.00 & 0.03 & 0.00 & \textbf{0.10} & 0.00 & 0.00 \\
\midrule
\multirow{7}{*}{\textbf{Omitted}} 
& ATE & 0.95 & 0.00 & \textbf{0.05} & 0.00 & 0.00 & 0.00 & 0.00 \\
& CDE & 0.97 & 0.00 & \textbf{0.03} & 0.00 & 0.00 & 0.00 & 0.00 \\
& ETT & 0.86 & 0.05 & \textbf{0.09} & 0.00 & 0.00 & 0.00 & 0.00 \\
& NDE & 0.93 & 0.00 & \textbf{0.07} & 0.00 & 0.00 & 0.00 & 0.00 \\
& NIE & 0.84 & 0.00 & \textbf{0.13} & 0.01 & 0.02 & 0.00 & 0.00 \\
& PN  & 0.92 & 0.00 & 0.00 & 0.00 & \textbf{0.08} & 0.00 & 0.00 \\
& PS  & 0.91 & 0.00 & 0.00 & 0.00 & \textbf{0.09} & 0.00 & 0.00 \\
\midrule
\multirow{7}{*}{\textbf{Redundant}} 
& ATE & 0.96 & 0.00 & \textbf{0.04} & 0.00 & 0.00 & 0.00 & 0.00 \\
& CDE & 0.98 & 0.00 & \textbf{0.02} & 0.00 & 0.00 & 0.00 & 0.00 \\
& ETT & 0.96 & 0.01 & \textbf{0.03} & 0.00 & 0.00 & 0.00 & 0.00 \\
& NDE & 0.86 & 0.00 & \textbf{0.11} & 0.01 & 0.02 & 0.00 & 0.00 \\
& NIE & 0.89 & 0.00 & \textbf{0.04} & 0.03 & \textbf{0.04} & 0.00 & 0.00 \\
& PN  & 0.95 & 0.00 & 0.01 & 0.00 & \textbf{0.04} & 0.00 & 0.00 \\
& PS  & 0.91 & 0.00 & 0.01 & 0.00 & \textbf{0.07} & 0.01 & 0.00 \\
\bottomrule
\end{tabular}
\centering
\caption{GRPO failure distribution across different test sets}
\label{tab:grpo_failure}
\end{table*}

\begin{table*}[!ht]
\small
\centering
\begin{tabular}{llccccccc}
\toprule
\textbf{Test set} & \textbf{Type} & \textbf{Correct} & \textbf{Task-ID} & \textbf{SCM-ID} & \textbf{Formula-ID} & \textbf{Formula-App} & \textbf{Num-Comp} & \textbf{Unexpected} \\
\midrule
\multirow{7}{*}{\textbf{CaLM}} 
& ATE & 0.84 & 0.01 & \textbf{0.08} & 0.00 & 0.00 & 0.01 & 0.06 \\
& CDE & 0.83 & 0.00 & \textbf{0.13} & 0.00 & 0.00 & 0.00 & 0.04 \\
& ETT & 0.51 & 0.17 & \textbf{0.29} & 0.03 & 0.00 & 0.00 & 0.00 \\
& NDE & 0.56 & 0.00 & \textbf{0.43} & 0.00 & 0.01 & 0.00 & 0.00 \\
& NIE & 0.64 & 0.00 & \textbf{0.31} & 0.01 & 0.03 & 0.01 & 0.00 \\
& PN  & 0.80 & 0.00 & 0.03 & 0.00 & \textbf{0.09} & 0.00 & 0.08 \\
& PS  & 0.73 & 0.00 & 0.07 & 0.00 & \textbf{0.12} & 0.02 & 0.06 \\
\midrule
\multirow{7}{*}{\textbf{Omitted}} 
& ATE & 0.43 & 0.03 & 0.23 & 0.00 & 0.00 & 0.00 & \textbf{0.31} \\
& CDE & 0.48 & 0.00 & 0.00 & 0.00 & 0.00 & 0.00 & \textbf{0.52} \\
& ETT & 0.27 & 0.05 & \textbf{0.68} & 0.00 & 0.00 & 0.00 & 0.00 \\
& NDE & 0.26 & 0.00 & \textbf{0.73} & 0.00 & 0.00 & 0.01 & 0.00 \\
& NIE & 0.42 & 0.00 & \textbf{0.49} & 0.00 & 0.04 & 0.01 & 0.04 \\
& PN  & 0.65 & 0.00 & 0.07 & 0.01 & \textbf{0.17} & 0.00 & 0.10 \\
& PS  & 0.53 & 0.00 & 0.07 & 0.01 & \textbf{0.29} & 0.01 & 0.09 \\
\midrule
\multirow{7}{*}{\textbf{Redundant}} 
& ATE & 0.81 & 0.00 & 0.05 & 0.00 & 0.00 & 0.01 & \textbf{0.13} \\
& CDE & 0.85 & 0.00 & 0.02 & 0.00 & 0.00 & 0.01 & \textbf{0.12} \\
& ETT & 0.29 & 0.04 & 0.19 & 0.01 & 0.00 & 0.00 & \textbf{0.47} \\
& NDE & 0.53 & 0.00 & 0.11 & 0.02 & 0.00 & 0.00 & \textbf{0.34} \\
& NIE & 0.67 & 0.00 & 0.07 & 0.00 & 0.01 & 0.00 & \textbf{0.25} \\
& PN  & 0.88 & 0.00 & 0.01 & 0.00 & 0.01 & 0.00 & \textbf{0.10} \\
& PS  & 0.76 & 0.00 & 0.02 & 0.00 & 0.04 & 0.00 & \textbf{0.18} \\
\bottomrule
\end{tabular}
\caption{SFT failure distribution across different test sets}
\label{tab:sft_failure}
\end{table*}

\begin{table*}[!ht]
\small
\centering
\begin{tabular}{llccccccc}
\toprule
\textbf{Test set} & \textbf{Type} & \textbf{Correct} & \textbf{Task-ID} & \textbf{SCM-ID} & \textbf{Formula-ID} & \textbf{Formula-App} & \textbf{Num-Comp} & \textbf{Unexpected} \\
\midrule
\multirow{7}{*}{\textbf{CaLM}} 
& ATE & 0.70 & 0.01 & 0.01 & 0.00 & 0.00 & 0.00 & \textbf{0.28} \\
& CDE & 0.53 & 0.00 & 0.00 & 0.00 & 0.00 & 0.00 & \textbf{0.47} \\
& ETT & 0.12 & 0.13 & \textbf{0.62} & 0.07 & 0.04 & 0.00 & 0.02 \\
& NDE & 0.30 & 0.01 & \textbf{0.54} & 0.01 & 0.00 & 0.00 & 0.14 \\
& NIE & 0.41 & 0.00 & 0.03 & 0.01 & 0.02 & 0.00 & \textbf{0.53} \\
& PN  & 0.82 & 0.00 & 0.00 & 0.00 & 0.00 & 0.00 & \textbf{0.18} \\
& PS  & 0.72 & 0.00 & 0.00 & 0.00 & 0.00 & 0.00 & \textbf{0.28} \\
\midrule
\multirow{7}{*}{\textbf{Omitted}} 
& ATE & 0.29 & 0.00 & 0.00 & 0.00 & 0.00 & 0.00 & \textbf{0.71} \\
& CDE & 0.30 & 0.00 & 0.00 & 0.00 & 0.00 & 0.00 & \textbf{0.70} \\
& ETT & 0.13 & 0.01 & 0.00 & 0.00 & 0.00 & 0.00 & \textbf{0.86} \\
& NDE & 0.16 & 0.00 & 0.00 & 0.00 & 0.00 & 0.00 & \textbf{0.84} \\
& NIE & 0.19 & 0.00 & 0.00 & 0.00 & 0.00 & 0.00 & \textbf{0.81} \\
& PN  & 0.60 & 0.00 & 0.00 & 0.00 & 0.00 & 0.00 & \textbf{0.40} \\
& PS  & 0.56 & 0.00 & 0.00 & 0.00 & 0.00 & 0.00 & \textbf{0.44} \\
\midrule
\multirow{7}{*}{\textbf{Redundant}} 
& ATE & 0.65 & 0.00 & 0.00 & 0.00 & 0.00 & 0.00 & \textbf{0.35} \\
& CDE & 0.60 & 0.00 & 0.00 & 0.00 & 0.00 & 0.00 & \textbf{0.40} \\
& ETT & 0.10 & 0.00 & 0.00 & 0.00 & 0.00 & 0.00 & \textbf{0.90} \\
& NDE & 0.28 & 0.00 & 0.00 & 0.00 & 0.00 & 0.00 & \textbf{0.72} \\
& NIE & 0.58 & 0.00 & 0.00 & 0.00 & 0.00 & 0.00 & \textbf{0.42} \\
& PN  & 0.70 & 0.00 & 0.00 & 0.00 & 0.00 & 0.00 & \textbf{0.30} \\
& PS  & 0.60 & 0.00 & 0.00 & 0.00 & 0.00 & 0.00 & \textbf{0.40} \\
\bottomrule
\end{tabular}
\caption{Cold Start Model failure distribution across different test sets}
\label{tab:cold_start_failure}
\end{table*}

\section{Comparison with Other Causal Reasoning Benchmark}
To show the difference between our benchmark \sys and other causal reasoning benchmarks, we compare \sys with CLadder \citep{jin2023cladder}, CaLM \citep{chen2024causal} and CLEAR \citep{chen-etal-2024-clear} and the result is shown in Table \ref{tab:causal_benchmarks_scaled}.
\begin{table*}[!ht]
    \centering

    \resizebox{\textwidth}{!}{
        \begin{tabular}{ccccccc}
            \hline
            \textbf{Benchmark} & \textbf{Training Dataset} & \textbf{Numerical Input} & \textbf{Rationale} & \textbf{Generalization Test} & \textbf{Internalization Test} & \textbf{Robustness Test} \\
            \hline
            CLadder & \textcolor{red}{\ding{55}} & \textcolor{green}{\ding{52}} & \textcolor{green}{\ding{52}} & \textcolor{red}{\ding{55}} & \textcolor{red}{\ding{55}} & \textcolor{red}{\ding{55}} \\
  
            CaLM & \textcolor{red}{\ding{55}} & \textcolor{green}{\ding{52}} & \textcolor{green}{\ding{52}} & \textcolor{red}{\ding{55}} & \textcolor{red}{\ding{55}} & \textcolor{red}{\ding{55}} \\
         
            CLEAR & \textcolor{red}{\ding{55}} & \textcolor{red}{\ding{55}} & \textcolor{red}{\ding{55}} & \textcolor{red}{\ding{55}} & \textcolor{red}{\ding{55}} & \textcolor{red}{\ding{55}} \\
        
            \sys & \textcolor{green}{\ding{52}} & \textcolor{green}{\ding{52}} & \textcolor{green}{\ding{52}} & \textcolor{green}{\ding{52}} & \textcolor{green}{\ding{52}} & \textcolor{green}{\ding{52}} \\
            \hline
        \end{tabular}
    } 
    \caption{ Comparison of causal reasoning benchmarks.}
    \label{tab:causal_benchmarks_scaled}
\end{table*}

\section{\sys Dataset Statistics}
To demonstrate the question diversity and entity variety of our training dataset and testing dataset, we list its critical characteristics in Table \ref{tab:training_set_stats} and \ref{tab:testing_set_stats}, where probability count refers to the number of distinct probabilities that appear in each question.

\begin{table*}[!ht]
    \centering
    \fontsize{9}{12}\selectfont
    {
    \begin{tabular}{ccccc}
        \hline
         \textbf{Maximal Node Number} & \textbf{Maximal Probability Count} & \textbf{Average Probability Count} & \textbf{Question Type} & \textbf{Sample} \\
        \hline
         10 & 12 & 3.84 & 7 & 17500 \\
        \hline
    \end{tabular}}
    \caption{Characteristics of the \sys training dataset.}
    \label{tab:training_set_stats}
\end{table*}

\begin{table*}[!ht]
    \centering
    \fontsize{9}{12}\selectfont
    {
    \begin{tabular}{ccccc}
        \hline
         \textbf{Dataset}& \textbf{Maximal Node Number} & \textbf{Average Probability Count} & \textbf{Question Type} & \textbf{Sample} \\
        \hline
         \hard & 8 & 6.41 & 7 & 700 \\
         \lack & 8 & 2.08 & 7 & 700 \\
         \tf & 5 & 3.11 & 7 & 700 \\
         \redun & 8 & 5.71 & 7 & 700 \\
         \reph & 5 & 3.11 & 7 & 700 \\
        \hline
    \end{tabular}}
    \caption{Characteristics of the \sys testing dataset.}
    \label{tab:testing_set_stats}
\end{table*}

\section{GRPO Model Performance on Other Causal Benchmark}
To further demonstrate the generalization capability of GRPO model, we evaluate the performance of GRPO model and its base model, \base~on other causal reasoning benchmarks, including CLadder, CaLM, CLEAR and CausalProbe-E \citep{chi2024unveiling}. As shown in Table \ref{tab:causal_performance}, the result demonstrates impressive generalization capability of GRPO model.

\begin{table*}[!ht]
    \centering
    \fontsize{9}{12}\selectfont
    \begin{tabular}{cccc}
        \hline
        \textbf{Benchmark} & \textbf{GRPO} & \textbf{Base} & \textbf{Best Performance Reported in the Original Paper} \\
        \hline
        CaLM (Mathematical) & 0.935&  0.272&  0.270 (GPT-3.5-Turbo) \\
  
        CLadder & 0.719 & 0.577&  0.704 (GPT-4 + CausalCOT) \\
 
        CLEAR & 0.472 & 0.433& 0.605 (GPT-4) \\
 
        CausalProbe-E & 0.805&  0.803 &0.750 (Claude 3.5 Opus)\\
        \hline
    \end{tabular}
    \caption{ Performance on other causal reasoning benchmarks.}
    \label{tab:causal_performance}
\end{table*}

\section{Post-trained Model's Reasoning Process}
\begin{figure*}[!ht]
    \centering
    \includegraphics[width=0.9\linewidth]{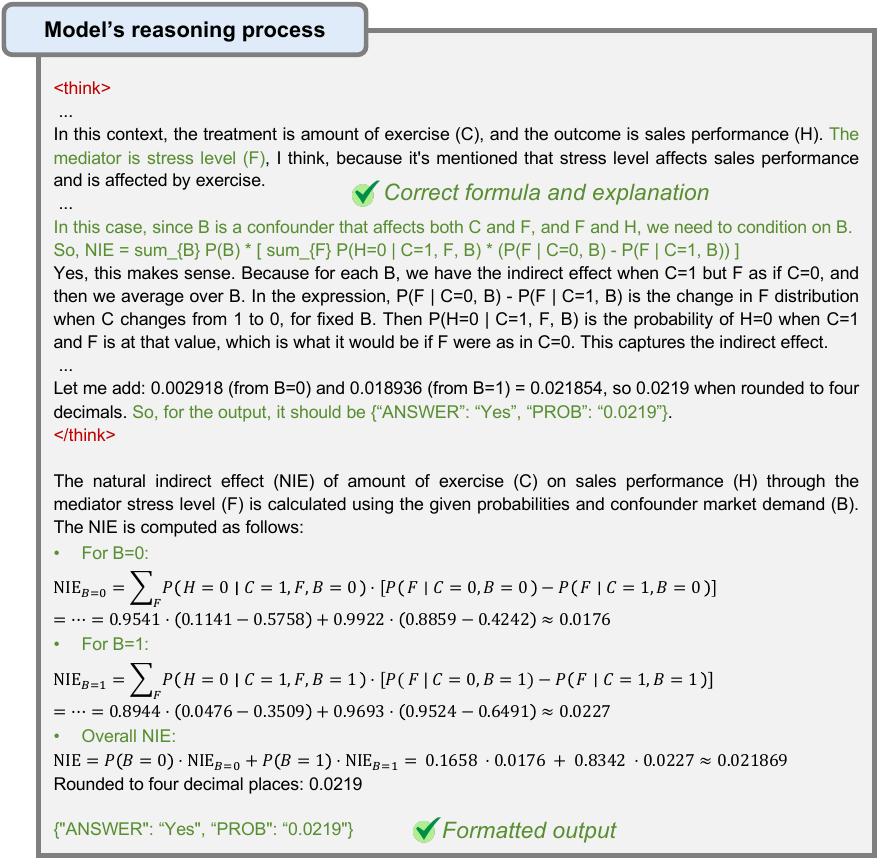}
    \caption{An example for GRPO model's reasoning process.}
    \label{fig:grpo_reasoning}
\end{figure*}

Figure \ref{fig:grpo_reasoning} demonstrates an example of how the GRPO model reason.

%% file: sections/2_preliminary.tex
\section{Preliminary}
\subsection{Structural Causal Model}
Structural Causal Model (SCM) is a way to describe causal-related variables and how they interact with each other \citep{pearl2016causal}. An SCM is a triple, represented as $\mathbf{M} = \{\mathbf{U}, \mathbf{V}, \mathbf{F}\}$. $\mathbf{U}$ denotes a set of exogenous variables, whose causes are outside the model. $\mathbf{V}$ denotes a set of endogenous variables, whose values are determined by variables within the model, namely the variables in $\mathbf{V}$ and $\mathbf{U}$. $\mathbf{F}$ denotes a set of functions, which specify how the value of each endogenous variable is determined. Their general form is $X=f_X({PA}_X,U_X)$, where $X$ is a variable in $\mathbf{V}$, $U_X$ is  a variable in $\mathbf{U}$, and ${PA}_X$ is a set of variables in $\mathbf{V}$, which have the direct effect to $X$.

An SCM can also be visualized as a directed acyclic graph (DAG) $G$. Its nodes represent the variables in $\mathbf{V}$. There is a directed edge from a variable $Y$ to $X$ in the $G$ if and only if $Y$ is a member of the $PA_X$ set for the function  $X=f_X({PA}_X,U_X)$.

\subsection{Intervention}
Intervention aims to answer the question of ``if'' (``\emph{If I lower the selling price now, will the sales increase?}''). It can be formally defined with the \emph{do-operator}, $P(Y=y \mid\textbf{do}(X=x))$. In the SCM, $\textbf{do}(X=x)$ means replacing the structural equation $X=f_X(PA_X,U_X)$ with $X=x$. 

A confounder is a variable that meets all three of these criteria: (1) It is a cause of the outcome, (2) It is associated with the treatment, (3) It is not a consequence of the treatment. If there are confounders for the outcome $Y$ and the treatment $X$, they will create a spurious correlation between the outcome and the treatment, namely $P(Y=y \mid \textbf{do}(X=x)) \neq P(Y=y \mid X=x)$. To identify the intervention effect with observational data, we can use the Backdoor Criterion. 

The backdoor criterion is defined as ``Given an ordered pair of variables $(X,Y)$ in a DAG $G$, a set of variables $Z$ satisfies the backdoor criterion relative to $(X,Y)$ if no node in $Z$ is a descendant of $X$, and $Z$ blocks every path between $X$ and $Y$ that contains an arrow into \textbf{$X$}'' \citep{pearl2016causal}. If a set of variables $Z$ satisfies this criterion for $X$ and $Y$, we can get the following formula:
\begin{align}
    &P(Y=y \mid \textbf{do}(X=x)) = \nonumber\\
    &\sum_{z} P(Y=y \mid X=x, Z=z) P(Z=z).
\end{align}


\subsection{Counterfactual}

Counterfactual reasoning addresses the ``what-if" question by estimating the values of variables under hypothetical interventions that differ from observed conditions. Formally, a counterfactual problem is often expressed as  $P(Y_{X=x}=y \mid X=x^{\prime},Y=y^{\prime})$. In this expression, $X=x^{\prime}$ and $Y=y^{\prime}$ represent the observed data, while $Y_{X=x}$ denotes the value of Y had the intervention X=x occurred. In a typical counterfactual calculation process, we use these observations to estimate the probability distributions (or the exact values) of exogenous variables. The process typically involves using observed data to estimate the probability distributions of exogenous variables within a causal model. Subsequently, an intervention is applied to the SCM, such as $\textbf{do}(X=x)$, to derive the final counterfactual outcomes. The connection between counterfactual inference and intervention inference is that $P(Y=y \mid \textbf{do}(X=x),Z=z)=P(Y_{X=x}=y \mid Z_{X=x}=z)$.

To identify and compute the counterfactual probability $P(Y_{X=x}=y)$ directly from empirical data, we can employ a theorem known as the  Counterfactual Interpretation of Backdoor \citep{pearl2016causal}. It states that if a set of variables, $Z$, satisfies the backdoor condition with respect to the causal relationship from $X$ to $Y$, then for any value $x$, the counterfactual outcome $Y_{X=x}$ is conditionally independent of the actual treatment $X$ given $Z$. This key property is formally expressed as:
\begin{align}
&P(Y_{X=x}=y \mid X=x^{\prime},Z=z)=\nonumber\\
&P(Y_{X=x}=y \mid Z=z).  
\end{align}


\subsection{Causal Inference Tasks}
In this paper, we delineate the following seven key causal inference tasks \citep{pearl2009causality,pearl2018book}:
(1) \textbf{Average Treatment Effect (ATE).} The expected difference in outcomes had everyone received treatment versus had everyone received no treatment.
(2) \textbf{Controlled Direct Effect (CDE).} The expected difference in outcomes had everyone received treatment versus had everyone received no treatment, while holding the mediator variable at a specific level.
(3) \textbf{Effect of the Treatment on the Treated (ETT).} The expected difference in outcomes for the subpopulation that actually received the treatment.
(4) \textbf{Natural Direct Effect (NDE).} The effect of the treatment on the outcome, with the mediator set to the value it would naturally take in the absence of treatment.
(5) \textbf{Natural Indirect Effect (NIE).} The effect on the outcome transmitted solely through the mediator, when the treatment is changed from no treatment to treatment.
(6) \textbf{Probability of Necessity (PN).} The probability that the absence of treatment was a necessary condition for the outcome to be absent, given that treatment was received and the outcome occurred.
(7) \textbf{Probability of Sufficiency (PS).} The probability that receiving treatment was a sufficient condition for the outcome to occur, given that no treatment was received and the outcome did not occur.

%% file: acl2026_conference.bib
@inproceedings{lewkowycz2022solving,
  title={Solving quantitative reasoning problems with language models},
  author={Lewkowycz, Aitor and Andreassen, Anders and Dohan, David and Dyer, Ethan and Michalewski, Henryk and Ramasesh, Vinay and Slone, Ambrose and Anil, Cem and Schlag, Imanol and Gutman-Solo, Theo and others},
  booktitle={Advances in neural information processing systems},
  year={2022}
}

@inproceedings{hendrycksmath2021,
  title={Measuring Mathematical Problem Solving With the MATH Dataset},
  author={Dan Hendrycks and Collin Burns and Saurav Kadavath and Akul Arora and Steven Basart and Eric Tang and Dawn Song and Jacob Steinhardt},
  booktitle={NeurIPS},
  year={2021}
}

@article{yu2025dapo,
  title={Dapo: An open-source llm reinforcement learning system at scale},
  author={Yu, Qiying and Zhang, Zheng and Zhu, Ruofei and Yuan, Yufeng and Zuo, Xiaochen and Yue, Yu and Dai, Weinan and Fan, Tiantian and Liu, Gaohong and Liu, Lingjun and others},
  journal={arXiv preprint arXiv:2503.14476},
  year={2025}
}

@inproceedings{hu2022lora,
  title={Lora: Low-rank adaptation of large language models.},
  author={Hu, Edward J and Shen, Yelong and Wallis, Phillip and Allen-Zhu, Zeyuan and Li, Yuanzhi and Wang, Shean and Wang, Lu and Chen, Weizhu and others},
  booktitle={ICLR},
  year={2022}
}

@article{qwen3,
    title={Qwen3 Technical Report}, 
    author={An Yang and Anfeng Li and Baosong Yang and Beichen Zhang and Binyuan Hui and Bo Zheng and Bowen Yu and Chang Gao and Chengen Huang and Chenxu Lv and Chujie Zheng and Dayiheng Liu and Fan Zhou and Fei Huang and Feng Hu and Hao Ge and Haoran Wei and Huan Lin and Jialong Tang and Jian Yang and Jianhong Tu and Jianwei Zhang and Jianxin Yang and Jiaxi Yang and Jing Zhou and Jingren Zhou and Junyang Lin and Kai Dang and Keqin Bao and Kexin Yang and Le Yu and Lianghao Deng and Mei Li and Mingfeng Xue and Mingze Li and Pei Zhang and Peng Wang and Qin Zhu and Rui Men and Ruize Gao and Shixuan Liu and Shuang Luo and Tianhao Li and Tianyi Tang and Wenbiao Yin and Xingzhang Ren and Xinyu Wang and Xinyu Zhang and Xuancheng Ren and Yang Fan and Yang Su and Yichang Zhang and Yinger Zhang and Yu Wan and Yuqiong Liu and Zekun Wang and Zeyu Cui and Zhenru Zhang and Zhipeng Zhou and Zihan Qiu},
    journal = {arXiv preprint arXiv:2505.09388},
    year={2025}
}

@inproceedings{gao2023chatgpt,
  title={Is chatgpt a good causal reasoner? a comprehensive evaluation},
  author={Gao, Jinglong and Ding, Xiao and Qin, Bing and Liu, Ting},
  booktitle={EMNLP},
  year={2023}
}

@book{woodward2005making,
  title={Making things happen: A theory of causal explanation},
  author={Woodward, James},
  year={2005},
  publisher={Oxford university press}
}

@book{bunge2017causality,
  title={Causality and modern science},
  author={Bunge, Mario},
  year={2017},
  publisher={Routledge}
}

@inproceedings{chen-etal-2024-clear,
    title = "{CLEAR}: Can Language Models Really Understand Causal Graphs?",
    author = "Chen, Sirui  and
      Xu, Mengying  and
      Wang, Kun  and
      Zeng, Xingyu  and
      Zhao, Rui  and
      Zhao, Shengjie  and
      Lu, Chaochao",
    booktitle = "EMNLP",
    year = "2024",
}

@book{pearl2018book,
  title={The book of why: the new science of cause and effect},
  author={Pearl, Judea and Mackenzie, Dana},
  year={2018},
  publisher={Basic books}
}

@inproceedings{shpitser2006identification,
  title={Identification of joint interventional distributions in recursive semi-Markovian causal models},
  author={Shpitser, Ilya and Pearl, Judea},
  booktitle={AAAI},
  year={2006}
}

@book{sloman2009causal,
  title={Causal models: How people think about the world and its alternatives},
  author={Sloman, Steven and Sloman, Steven A},
  year={2009},
  publisher={Oxford University Press}
}

@article{zevcevic2023causal,
  title={Causal parrots: Large language models may talk causality but are not causal},
  author={Ze{\v{c}}evi{\'c}, Matej and Willig, Moritz and Dhami, Devendra Singh and Kersting, Kristian},
  journal={Transactions on Machine Learning Research},
  year={2023}
}

@article{schulman2017proximal,
  title={Proximal policy optimization algorithms},
  author={Schulman, John and Wolski, Filip and Dhariwal, Prafulla and Radford, Alec and Klimov, Oleg},
  journal={arXiv preprint arXiv:1707.06347},
  year={2017}
}

@inproceedings{ouyang2022training,
  title={Training language models to follow instructions with human feedback},
  author={Ouyang, Long and Wu, Jeffrey and Jiang, Xu and Almeida, Diogo and Wainwright, Carroll and Mishkin, Pamela and Zhang, Chong and Agarwal, Sandhini and Slama, Katarina and Ray, Alex and others},
  booktitle={NeurIPS},
  year={2022}
}

@inproceedings{ethayarajh2024kto,
  title={Kto: Model alignment as prospect theoretic optimization},
  author={Ethayarajh, Kawin and Xu, Winnie and Muennighoff, Niklas and Jurafsky, Dan and Kiela, Douwe},
  booktitle={ICML},
  year={2024}
}

@inproceedings{rafailov2023direct,
  title={Direct preference optimization: Your language model is secretly a reward model},
  author={Rafailov, Rafael and Sharma, Archit and Mitchell, Eric and Manning, Christopher D and Ermon, Stefano and Finn, Chelsea},
  booktitle={NeurIPS},
  year={2023}
}

@inproceedings{
wei2022finetuned,
title={Finetuned Language Models are Zero-Shot Learners},
author={Jason Wei and Maarten Bosma and Vincent Zhao and Kelvin Guu and Adams Wei Yu and Brian Lester and Nan Du and Andrew M. Dai and Quoc V Le},
booktitle={ICLR},
year={2022},
}

@article{shao2024deepseekmath,
  title={Deepseekmath: Pushing the limits of mathematical reasoning in open language models},
  author={Shao, Zhihong and Wang, Peiyi and Zhu, Qihao and Xu, Runxin and Song, Junxiao and Bi, Xiao and Zhang, Haowei and Zhang, Mingchuan and Li, YK and Wu, Yang and others},
  journal={arXiv preprint arXiv:2402.03300},
  year={2024}
}

@misc{econml,
  author={Battocchi, Keith and Dillon, Eleanor and Hei, Maggie and Lewis, Greg and Oka, Paul and Oprescu, Miruna and Syrgkanis, Vasilis },
  title={{EconML}: {A Python Package for ML-Based Heterogeneous Treatment Effects Estimation}},
  howpublished={https://github.com/py-why/EconML},
  note={Version 0.x},
  year={2019}
}

@misc{chen2020causalml, title={CausalML: Python Package for Causal Machine Learning}, author={Huigang Chen and Totte Harinen and Jeong-Yoon Lee and Mike Yung and Zhenyu Zhao}, year={2020}, eprint={2002.11631}, archivePrefix={arXiv}, primaryClass={cs.CY} }

@article{dowhy,
  title={DoWhy: An End-to-End Library for Causal Inference},
  author={Sharma, Amit and Kiciman, Emre},
  journal={arXiv preprint arXiv:2011.04216},
  year={2020}
}

@article{rubin1974estimating,
  title={Estimating causal effects of treatments in randomized and nonrandomized studies.},
  author={Rubin, Donald B},
  journal={Journal of educational Psychology},
  year={1974},
  publisher={American Psychological Association}
}

@InProceedings{pmlr-v6-pearl10a,
  title = 	 {Causal Inference},
  author = 	 {Pearl, Judea},
  booktitle = 	 {NeurIPS},
  year = 	 {2010},
}

@book{pearl2016causal,
  title={Causal inference in statistics: A primer},
  author={Pearl, Judea and Glymour, Madelyn and Jewell, Nicholas P},
  year={2016},
  publisher={John Wiley \& Sons}
}

@inproceedings{quan2024verification,
  title={Verification and Refinement of Natural Language Explanations through LLM-Symbolic Theorem Proving},
  author={Quan, Xin and Valentino, Marco and Dennis, Louise and Freitas, Andr{\'e}},
  booktitle={EMNLP},
  year={2024}
}

@inproceedings{
luo2025wizardmath,
title={WizardMath: Empowering Mathematical Reasoning for Large Language Models via Reinforced Evol-Instruct},
author={Haipeng Luo and Qingfeng Sun and Can Xu and Pu Zhao and Jian-Guang Lou and Chongyang Tao and Xiubo Geng and Qingwei Lin and Shifeng Chen and Yansong Tang and Dongmei Zhang},
booktitle={ICLR},
year={2025}
}

@inproceedings{wang2024math,
  title={Math-Shepherd: Verify and Reinforce LLMs Step-by-step without Human Annotations},
  author={Wang, Peiyi and Li, Lei and Shao, Zhihong and Xu, Runxin and Dai, Damai and Li, Yifei and Chen, Deli and Wu, Yu and Sui, Zhifang},
  booktitle={ACL},
  year={2024}
}

@inproceedings{
guan2025rstarmath,
title={rStar-Math: Small {LLM}s Can Master Math Reasoning with Self-Evolved Deep Thinking},
author={Xinyu Guan and Li Lyna Zhang and Yifei Liu and Ning Shang and Youran Sun and Yi Zhu and Fan Yang and Mao Yang},
booktitle={ICML},
year={2025},
}

@article{guo2025deepseek,
  title={Deepseek-r1: Incentivizing reasoning capability in llms via reinforcement learning},
  author={Guo, Daya and Yang, Dejian and Zhang, Haowei and Song, Junxiao and Zhang, Ruoyu and Xu, Runxin and Zhu, Qihao and Ma, Shirong and Wang, Peiyi and Bi, Xiao and others},
  journal={arXiv preprint arXiv:2501.12948},
  year={2025}
}

@inproceedings{
jin2024can,
title={Can Large Language Models Infer Causation from Correlation?},
author={Zhijing Jin and Jiarui Liu and Zhiheng LYU and Spencer Poff and Mrinmaya Sachan and Rada Mihalcea and Mona T. Diab and Bernhard Sch{\"o}lkopf},
booktitle={ICLR},
year={2024},
}

@inproceedings{nam2024using,
  title={Using an llm to help with code understanding},
  author={Nam, Daye and Macvean, Andrew and Hellendoorn, Vincent and Vasilescu, Bogdan and Myers, Brad},
  booktitle={ICSE},
  year={2024}
}

@article{rubin2005causal,
  title={Causal inference using potential outcomes: Design, modeling, decisions},
  author={Rubin, Donald B},
  journal={Journal of the American statistical Association},
  year={2005},
}

@misc{meta2024llama3,
  author = {Meta},
  title = {Model Cards of Llama 3.3},
  year = {2024},
  publisher = {Meta Github},
  journal = {Github Repository},
  howpublished = {\url{https://github.com/meta-llama/llama-models/blob/main/models/llama3_3/MODEL_CARD.md}}
}

@misc{amc2023math,
  author = {AMC},
  title = {AMC 2023},
  year = {2023},
  publisher = {Qwen Github},
  journal = {Github Repository},
  howpublished = {\url{https://github.com/QwenLM/Qwen2.5-Math/blob/main/evaluation/data/amc23/test.jsonl}}
}

@misc{deepmind2025gemini,
  author = {Deepmind},
  title = {Gemini 2.5 Pro},
  year = {2025},
  publisher = {Deepmind Webpage},
  journal = {Blog},
  howpublished = {\url{https://deepmind.google/models/gemini/pro/}}
}

@misc{openai2025o3,
  author = {OpenAI},
  title = {Introducing OpenAI o3 and o4-mini},
  year = {2025},
  publisher = {OpenAI Webpage},
  journal = {Blog},
  howpublished = {\url{https://openai.com/index/introducing-o3-and-o4-mini/}}
}

@inproceedings{jin2023cladder,
  title={Cladder: Assessing causal reasoning in language models},
  author={Jin, Zhijing and Chen, Yuen and Leeb, Felix and Gresele, Luigi and Kamal, Ojasv and Lyu, Zhiheng and Blin, Kevin and Gonzalez Adauto, Fernando and Kleiman-Weiner, Max and Sachan, Mrinmaya and others},
  booktitle={NeurIPS},
  year={2023}
}

@inproceedings{chen2024causal,
  title={Causal Evaluation of Language Models},
  author={Chen, Sirui and Peng, Bo and Chen, Meiqi and Wang, Ruiqi and Xu, Mengying and Zeng, Xingyu and Zhao, Rui and Zhao, Shengjie and Qiao, Yu and Lu, Chaochao},
  booktitle={CoRR},
  year={2024}
}

@inproceedings{chen-etal-2024-cello,
    title = "{CELLO}: Causal Evaluation of Large Vision-Language Models",
    author = "Chen, Meiqi  and
      Peng, Bo  and
      Zhang, Yan  and
      Lu, Chaochao",
    booktitle = "EMNLP",
    year = "2024",
}

@inproceedings{chi2024unveiling,
  title={Unveiling causal reasoning in large language models: Reality or mirage?},
  author={Chi, Haoang and Li, He and Yang, Wenjing and Liu, Feng and Lan, Long and Ren, Xiaoguang and Liu, Tongliang and Han, Bo},
  booktitle={NeurIPS},
  year={2024}
}

@book{pearl2009causality,
  title={Causality},
  author={Pearl, Judea},
  year={2009},
  publisher={Cambridge university press}
}
